\documentclass{article}

    \PassOptionsToPackage{numbers, compress}{natbib}
 \usepackage[preprint]{neurips_2026}

\usepackage[utf8]{inputenc} 
\usepackage[T1]{fontenc}    
\usepackage{hyperref}       
\usepackage{url}            
\usepackage{booktabs}       
\usepackage{amsfonts}       
\usepackage{nicefrac}       
\usepackage{microtype}      
\usepackage{xcolor}         

\usepackage{graphicx}
\usepackage{subcaption}
\usepackage{tikz}
\usepackage{amsmath}
\usepackage{dsfont}
\usepackage{longtable}
\usepackage{tabularx,colortbl}
\usepackage{multirow}
\usepackage{soul}

\title{StableTTA: Improving Vision Model Performance by Training-free Test-Time Adaptation Methods}

\author{%
  Zheng Li \\
  \small{Department of Computer Science}\\
  \small{New York Institute of Technology}\\
  \texttt{zli66@nyit.edu} \\
  \And
  Jerry Cheng \\
  \small{Department of Computer Science}\\
  \small{New York Institute of Technology}\\
  \texttt{jcheng18@nyit.edu} \\
  \AND
  Huanying Helen Gu \\
  \small{Department of Computer Science}\\
  \small{New York Institute of Technology}\\
  \texttt{hgu03@nyit.edu} \\
}

\begin{document}

\maketitle

\begin{abstract}
  Ensemble methods improve predictive performance but often incur high memory and computational costs. We identify an aggregation instability induced by nonlinear projection and voting operations. To address both efficiency challenges and this inconsistency, we propose StableTTA, a training-free test-time adaptation method with two variants.
  StableTTA-I targets coherent-batch inference settings, where temporally or semantically adjacent observations are likely to belong to the same class. Examples include burst photography, video streams, robotics perception, and industrial inspection.
  Under coherent-batch inference, StableTTA-I substantially improves prediction consistency and accuracy through variance-aware logit aggregation.
  StableTTA-II establishes feature-level cropping, enabling efficient logit aggregation with a single forward pass on a single model backbone.
  Experiments on ImageNet-1K across 71 models demonstrate that StableTTA-I consistently improves prediction accuracy under coherent-batch inference, while StableTTA-II provides lightweight and architecture-agnostic accuracy improvements with minimal computational overhead. These results suggest that inference-time semantic coherence and aggregation stability provide useful perspectives for improving practical test-time adaptation systems.
\end{abstract}

\section{Introduction}
\label{sec:Introduction}

Ensemble methods improve predictive robustness by aggregating multiple predictions \citep{hastie2009elements,breiman1996bagging}. 
However, their practical adoption is limited by substantial memory and computational overhead, since inference cost typically scales linearly with the number of experts. Test-time augmentation (TTA) partially alleviates this issue by applying multiple augmentations to a single model \citep{krizhevsky2012imagenet}, but still requires repeated forward passes and therefore remains computationally expensive.
In this work, we examine the limitations of ensemble methods and identify an inherent aggregation inconsistency between different aggregation strategies. While conventional image classification assumes independently sampled test examples, many real-world deployment scenarios exhibit strong local semantic consistency. For example, consecutive video frames, burst photographs, robotic observations, medical slices, and industrial inspection streams often contain correlated views of the same semantic object or category. In these settings, neighboring samples provide contextual information that is ignored by standard test-time augmentation methods. We refer to this setting as \textit{coherent-batch inference}, where semantically adjacent observations are processed jointly during inference. More generally, inference settings in visual recognition systems can be categorized according to the statistical relationships among test observations. While most existing image classification benchmarks and test-time adaptation methods assume independently sampled inputs, many practical deployment scenarios exhibit structured dependencies across observations.
To better position our setting, Table~\ref{tab:inference_regimes} summarizes several representative inference regimes commonly encountered in visual recognition systems. Most existing TTA methods are developed primarily for IID inference, whereas StableTTA-I explicitly exploits deployment-time semantic coherence during aggregation.

Motivated by these observations, we propose StableTTA, a training-free test-time adaptation framework consisting of two complementary methods.
StableTTA-I targets coherent batch inference and introduces a consensus-preserving logit-aggregation strategy for semantically correlated observations. Under coherent-batch inference, StableTTA-I substantially improves prediction consistency and accuracy.
StableTTA-II addresses the computational limitations of conventional ensemble methods by introducing feature-level cropping, enabling multiple predictions from a single backbone forward pass with only negligible additional computation in the lightweight classification head.

\begin{table}[t]
\centering
\caption{Representative inference regimes in visual recognition systems.}
\label{tab:inference_regimes}
\begin{tabular}{l l l}
\toprule
Regime & Assumption & Example Applications \\
\midrule
IID inference & Independent samples & Standard image classification \\
Coherent-batch inference & Neighboring samples correlated & Burst imaging, inspection \\
Multi-view inference & Multiple views of same object & 3D recognition \\
Sequential inference & Temporal continuity & Video and robotics \\
\bottomrule
\end{tabular}
\end{table}

\section{Related Work}
\label{sec:Related Work}
For image classification, model averaging is an ensemble strategy widely used to improve performance in general settings \citep{hastie2009elements}. As introduced by \citet{breiman1996bagging}, the outputs of multiple independently pretrained models are typically aggregated by majority voting on the predictions or averaging the predicted probabilities. In general,  although model averaging can provide modest performance gains, these improvements are often marginal relative to the substantial increase in resource requirements \citep{he2016deep,tan2019efficientnet}. Furthermore, both the total number of parameters and the computational cost (FLOPs) grow additively with the number of models. As a result, model averaging is typically less efficient than training a well-optimized single model~\citep{he2016deep}.

As a partial solution, TTA uses a single model to reduce the total number of model parameters. Specifically, it applies a set of data augmentations (e.g., flipping, cropping) to an input image and feeds the augmented images into the same model to produce multiple logits for aggregation \citep{krizhevsky2012imagenet,he2016deep}, while maintaining a fixed model size. However, it still incurs additional inference-time computation. For instance, ResNet \citep{he2016deep} and DenseNet \citep{huang2017densely} employ a 10-crop data augmentation during validation, leading to a tenfold increase in FLOPs. As a result, TTA typically yields only a modest improvement, often no more than 2\% in top-1 accuracy~\citep{shanmugam2021better,kim2020learning,li2025losstransform}.

Although TTA has been widely adopted, relatively few works have gained significant attention in top-tier venues, primarily because its key limitation—\textbf{the significant increase in computational cost}—remains unresolved \citep{shanmugam2025test}.

Historically, progress in large-scale visual recognition benchmarks such as ImageNet-1K (1.28M images) \citep{deng2009imagenet} has been characterized by incremental improvements, with successive model generations typically achieving only 1\%–2\% gains in top-1 accuracy \citep{krizhevsky2012imagenet,simonyan2014very,he2016deep,tan2019efficientnet,dosovitskiy2020image}. While some models report accuracies exceeding 85\% \citep{dosovitskiy2020image,liu2022convnet}, these gains are often enabled by scaling training data using larger external datasets such as ImageNet-21K (14M images) \citep{ridnik2021imagenet} and JFT-300M (300M images) \citep{mahajan2018exploring}, highlighting the role of data scale in achieving state-of-the-art performance.

\section{Background}
\label{sec:Background}
\begin{figure}
    \centering 
    \includegraphics[width=\columnwidth]{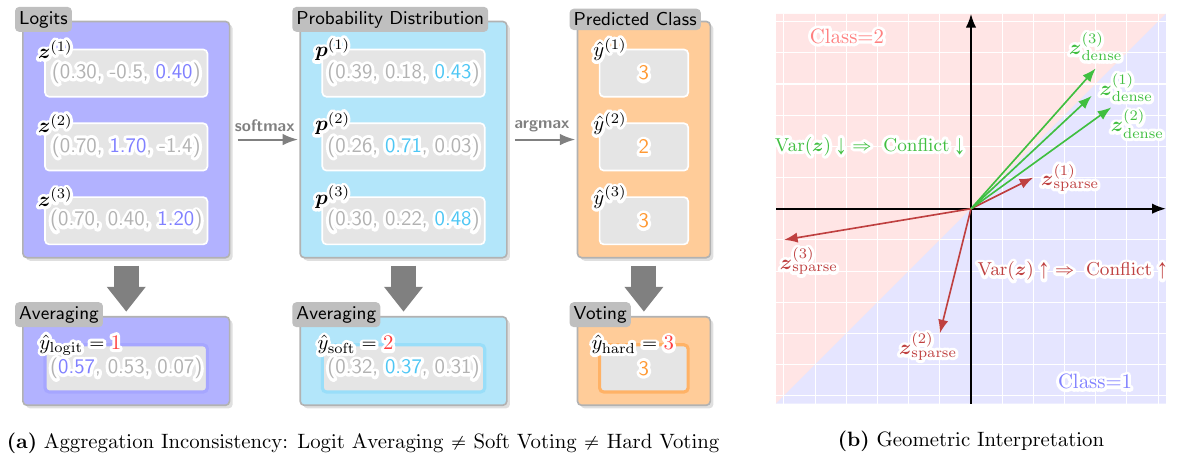} 
    \caption{
    (a) \textbf{Aggregation Inconsistency:} Given branch logits \(\{\boldsymbol{z}^{(i)} \mid i=1, 2, 3\}\), probabilities \(\boldsymbol{p}^{(i)}=\operatorname{softmax}(\boldsymbol{z}^{(i)})\), and predictions \(\hat{y}^{(i)}=\operatorname{argmax} \boldsymbol{p}^{(i)}\), different aggregation strategies may produce inconsistent results. Here, logit averaging predicts \(\hat{y}_{\text{logit}}=1\), soft voting predicts \(\hat{y}_{\text{soft}}=2\), and hard voting predicts \(\hat{y}_{\text{hard}}=3\).
    (a) \textbf{Explanation:} When \(({\boldsymbol{z}^{(1)}, \boldsymbol{z}^{(2)}, \dots})\) are sparsely distributed, inconsistency \(\hat{y}_{\text{logit}} \neq \hat{y}_{\text{soft}} \neq \hat{y}_{\text{hard}} \) are more likely due to the nonlinearity and non-bijectivity of softmax. In contrast, when the logits are more densely clustered, such inconsistencies are less likely.} 
    \label{fig:conflict}
\end{figure}

For a better understanding of the most commonly used ensemble methods, consider a \(C\)-class classification task. Let logit vector \(\boldsymbol{z}\) denote the outputs of a model \(f\) with parameters \(\boldsymbol{\theta}\) for input \(\boldsymbol{x}\):
\[
    \boldsymbol{z} = f(\boldsymbol{x}; \boldsymbol{\theta}) \in \mathbb{R}^C.
\]
After applying the softmax function, we obtain the probability distribution \(\boldsymbol{p}_{\boldsymbol{\theta}}(y \mid \boldsymbol{x})\) of class \(y\) over the \(C\) classes:
\[
    \boldsymbol{p}_{\boldsymbol{\theta}}(y=k \mid \boldsymbol{x}) = \frac{\exp(\boldsymbol{z}_k)}{\sum_{j=1}^C \exp(\boldsymbol{z}_j)}, \quad k=1, \dots, C.
\]
The predicted class is given by
\[
    \hat{y} = \operatorname*{arg\,max}_k \boldsymbol{z}_k = \operatorname*{arg\,max}_k \boldsymbol{p}_{\boldsymbol{\theta}} (y=k \mid \boldsymbol{x}).
\]
\textit{Hard voting}, \textit{soft voting}, and \textit{logit averaging} are common aggregation strategies in ensemble methods \cite{hastie2009elements,pedregosa2011scikit}. Among them, logit averaging is generally more effective for single-model ensembles (see Appendix \ref{app:Preliminary Concepts and Reproduction of Prior Work}, Table \ref{tab:survey_ours}). For an ensemble of \(N\) experts, the aggregation can be formulated as follows:
\begin{itemize}
    \item  Hard Voting (as illustrated by orange blocks in Fig.~\ref{fig:conflict}a):
    After all \(N\) votes are cast as \(\{\hat{y}^{(1)}, \hat{y}^{(2)}, \dots, \hat{y}^{(N)} \}\), the most favorable class is selected as:
    \[
        \hat{y}_{\text{hard}} = \operatorname*{arg\,max}_k \sum_{i=1}^{N} \mathds{1}_{\hat{y}^{(i)} = k}.
    \]
    \item Soft Voting (as illustrated by cyan blocks in Fig.~\ref{fig:conflict}a):
    After class probabilities from the \(N\) forward passes are averaged, the class with the highest mean probability is selected. Specifically, in the case of multi-model ensembles with models \( \{f(\cdot;\boldsymbol{\theta}^{(i)}) \mid i=1, \dots, N\} \), we have
    \[
        \hat{y}_{\text{soft}} = \operatorname*{arg\,max}_k \frac{1}{N} \sum_{i=1}^{N} \boldsymbol{p}_{\boldsymbol{\theta}^{(i)}}(y=k \mid \boldsymbol{x}).
    \]
    In the case of TTA with a single model \(f(\cdot;\boldsymbol{\theta})\) and augmentation policies \( \{\psi^{(i)} \mid i=1, \dots, N\} \), we have
    \[
        \hat{y}_{\text{soft}} = \operatorname*{arg\,max}_k \frac{1}{N} \sum_{i=1}^{N} \boldsymbol{p}_{\boldsymbol{\theta}}(y=k \mid \psi^{(i)}(\boldsymbol{x})).
    \]
    \item Logit Averaging (as illustrated by orange blocks in Fig.~\ref{fig:conflict}a):
    After logits are averaged across \(N\) forward passes, the class with the highest mean logit is selected as:
    \[
        \hat{y}_{\text{logit}} = \operatorname*{arg\,max}_k \frac{1}{N} \sum_{i=1}^{N} \boldsymbol{z}_k^{(i)}.
    \]
\end{itemize}

\paragraph{Limitations} Model averaging requires storing multiple models, and its inference-time cost equals the sum of all models. In contrast, TTA uses a single model but still incurs an N-fold increase in inference-time cost. Both methods yield only modest improvements. To better illustrate this, Fig. \ref{fig:background} in Appendix \ref{app:Preliminary Concepts and Reproduction of Prior Work} compares the pipelines of a single-model one forward pass, multi-model averaging, and TTA.

\paragraph{Aggregation Inconsistency} 
In ensemble methods, different aggregation strategies can yield inconsistent results. For example, as illustrated in Fig. \ref{fig:conflict}b,  given three logit vectors \((\boldsymbol{z}^{(1)}, \boldsymbol{z}^{(2)}, \boldsymbol{z}^{(3)})\), the aggregated predictions are inconsistent: 3 from hard voting, 2 from soft voting, and 1 from logit averaging. 
We also observe that when the logits \(\{ \boldsymbol{z}^{(1)}, \boldsymbol{z}^{(2)}, \dots \}\) are sparsely distributed, the nonlinearity and non-bijective nature of the softmax and indicator functions can amplify the differences among them. 
In contrast, when the logits are more densely clustered, different aggregation strategies are more likely to yield consistent predictions. 
We provide both theoretical analysis and empirical evidence in Appendix \ref{app:Jarque–Bera Test of StableTTA-I} and \ref{app:Monte Carlo Simulation of StableTTA-I} to support this observation. 
The empirical results are well fitted by our derived relationship between \(\operatorname{P}(\hat{y}_{\text{logit}} \neq \hat{y}_{\text{hard}})\) and \(\operatorname{Var}(\boldsymbol{z})\). In summary, reduced variance leads to more inconsistencies.

\section{Method}

\subsection{StableTTA-I}
\label{sec:StableTTA-I}

As discussed in Section~\ref{sec:Introduction}, StableTTA-I targets coherent-batch inference, where neighboring observations exhibit strong semantic correlation and therefore provide complementary predictive evidence during aggregation.
Formally, let \((\boldsymbol{x}_i, y_i)\) denote sequential observations. We assume there exists a locality scale \(\delta > 0\) such that
\[
P(y_i = y_j \mid |i-j| < \delta) \gg \frac{1}{C}
\]
where \(C\) denotes the number of classes. This assumption implies that neighboring samples are substantially more likely to share the same semantic category than under IID sampling.

Under this setting, aggregation stability becomes particularly important because predictions from neighboring observations and augmented samples are jointly combined during inference. As discussed in Section~\ref{sec:Background}, aggregation inconsistency arises because averaging in the logit, probability, and label spaces yields different behaviors after nonlinear projection via the softmax/argmax functions.

Motivated by this observation, we introduce a consensus-preserving logit post-processing operator called Non-Significant Suppression (NSS), which suppresses unstable low-confidence posterior modes prior to aggregation:
\begin{equation}
    \text{NSS}(\boldsymbol{z},K)
    :=
    \boldsymbol{z} \odot \mathbf{1}_{\operatorname{TopK}(\boldsymbol{z},K)} + \min(\boldsymbol{z}) \cdot (\mathbf{1} - \mathbf{1}_{\operatorname{TopK}(\boldsymbol{z},K)}), \quad K=1, \dots, C
    \label{eq:NSS}
\end{equation}
where, \( \mathbf{1} \) is the all-ones vector, \( \mathbf{1}_{\operatorname{TopK}(\boldsymbol{z},K)} \in \{0,1\}^C \) is the indicator vector of the Top-\(K\) indices (equal to 1 for indices in the Top-\(K\) set and 0 otherwise), and \(\odot\) denotes element-wise multiplication. In this work, we apply NSS to logit averaging and prove it reduces \(\text{Var}(\boldsymbol{z})\).

From the definition of NSS, we have:
\[
\mathrm{NSS}(\boldsymbol{z},K)_i = 
\begin{cases}
\boldsymbol{z}_i, & i \in \operatorname{TopK}(\boldsymbol{z},K), \\
\min(\boldsymbol{z}), & i\notin \operatorname{TopK}(\boldsymbol{z},K),
\end{cases}
\]
and 
\[
\frac{\partial}{\partial \boldsymbol{z}_j} \mathrm{NSS}(\boldsymbol{z},K)_i = 
\begin{cases}
1, & \text{~if~} i \in \operatorname{TopK}(\boldsymbol{z},K) \text{~and~} i = j, \\
0, & \text{~if~} i \in \operatorname{TopK}(\boldsymbol{z},K) \text{~and~} i \neq j, \\
1, & \text{~if~} i \notin \operatorname{TopK}(\boldsymbol{z},K) \text{~and~} j = \operatorname{arg\,min} \boldsymbol{z}, \\
0, & \text{~if~} i \notin \operatorname{TopK}(\boldsymbol{z},K) \text{~and~} j \neq \operatorname{arg\,min} \boldsymbol{z}.
\end{cases}
\]
Hence, almost everywhere,
\[
\left\| \nabla \mathrm{NSS}(\boldsymbol{z},K)_i \right\|^2 
 = \sum_{j=1}^C \left( \frac{\partial }{\partial \boldsymbol{z}_j} \mathrm{NSS}(\boldsymbol{z},K)_i \right)^2 
= 1.
\]
Our experiments in Appendix~\ref{app:Jarque–Bera Test of StableTTA-I} and \ref{app:Monte Carlo Simulation of StableTTA-I} show that, under our proposed augmentation policy and image sampling strategy, in most cases there exist \(\mu_i\) and \(\sigma\) such that \( \boldsymbol{z}_i \sim \mathcal{N}(\mu_i, \sigma^2)\). Applying the Gaussian Poincaré inequality, we obtain
\[
\operatorname{Var} (\operatorname{NSS}(\boldsymbol{z},K)_i) 
\le \mathbb{E} \left[ \sigma^2 \left\| \nabla \mathrm{NSS}(\boldsymbol{z},K)_i \right\|^2 \right] = \sigma^2 
= \operatorname{Var} (\boldsymbol{z}_i).
\]
This implies that \(\operatorname{NSS}\) reduces the variance of the logit vector $\boldsymbol{z}$, thereby further avoiding the prediction inconsistency described in Section~\ref{sec:Background}. Furthermore, it is important to note that
\[
\mathrm{NSS}(\boldsymbol{z},C) = \boldsymbol{z} \odot \mathbf{1} + \min(\boldsymbol{z}) \cdot (\mathbf{1} - \mathbf{1}) = \boldsymbol{z},
\]
which implies that when \(K=C\), StableTTA-I is downgraded to the conventional TTA method. 

To further reduce variance, we reformulate mixup \cite{zhang2018mixup} and CutMix \cite{yun2019cutmix} under the constraint that images within each batch mostly belong to the same class (see Appendix~\ref{app:Sequential Sampling and Data Augmentation Policies of StatbleTTA-I}). Although this image sampling strategy restricts the scope of applicability (e.g., requiring multiple captures of the same object for StableTTA-I), it yields substantial improvements in model performance.

In summary, StatleTTA-I uses the following representation to replace the logits \(\boldsymbol{z}\):
\[ 
\boldsymbol{z} \leftarrow \frac{1}{N} \sum_{i=1}^N \operatorname{NSS}(f(\psi^{(i)}(\boldsymbol{x}); \boldsymbol{\theta}),K).
\]
In Section \ref{sec:Evaluations of StableTTA-I}, we present an overview of the comparison experiments, ablation studies, and sensitivity analysis. A complete comparison with conventional TTA methods (under the same image sampling strategy as ours) is provided in Appendix \ref{app:Comparison Experiments of StableTTA-I}. In order to empirically demonstrate that NSS can mitigate the inconsistency, we provide detailed ablation studies in Appendix \ref{app:Ablation Study of StableTTA-I}, which show that, under the same sampling strategy and augmentation policy, applying NSS improves model performance. Additional results on sensitivity analysis are reported in Appendix \ref{app:Sensitivity Analysis of StableTTA-I}. Appendix \ref{app:Means and Standard Deviations of Repeated Experimental Evaluation of StableTTA-I} presents the statistical significance of the experimental results.

\subsection{StableTTA-II}
\label{sec:StableTTA-II}
In this section, we present StableTTA-II, designed to address the limitations discussed in Section \ref{sec:Background}.

\begin{figure}
    \centering 
    \includegraphics[width=\columnwidth]{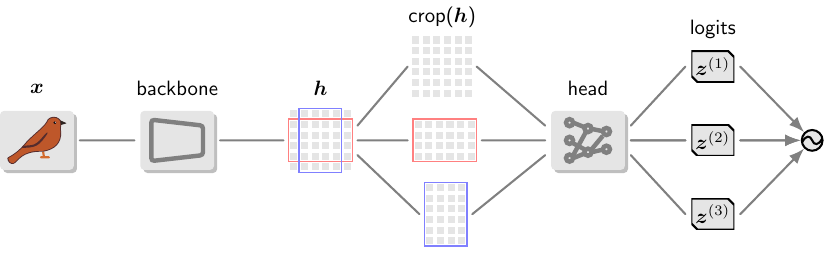} 
    \caption{Illustration of StableTTA-II. Given an input image \(\boldsymbol{x}\), the backbone produces a feature representation \(\boldsymbol{h}\). Multiple deterministic crops are sampled at the feature level and processed by a shared head to obtain multiple logits \( \{\boldsymbol{z}^{(i)}\} \), which are aggregated to form the final prediction.} 
    \label{fig:pipeline}
\end{figure}

To ensure unchanged memory usage and minimal additional computational cost, we constrain our method to a single model and a single forward pass for logit aggregation. This contrasts with conventional test-time augmentation (TTA), which requires multiple forward passes and incurs significantly higher computational cost. To achieve this goal, we do not apply input augmentations to generate multiple logits, as doing so would scale the total FLOPs (see Appendix~\ref{app:Preliminary Concepts and Reproduction of Prior Work}, Fig.~\ref{fig:background}). Instead, we apply feature-level cropping to produce multiple logits (see Fig.~\ref{fig:pipeline}).

In general, most of the computational cost (FLOPs) lies in the model backbone. Although the head is executed multiple times, the overall FLOPs remain nearly unchanged. For example, consider the ResNet-50 architecture with a single \(224\times224\) input resolution: the backbone consists of conv + bn/relu/maxpool + residual blocks 1–4, totaling 4.087 GFLOPs, while the head consists of avgpool + flatten + linear, totaling 0.002 GFLOPs. Accordingly, StableTTA-II achieves a single forward pass on the main backbone while introducing a negligible increase in overall computational cost.

We decompose the model \(f\) into a feature extractor (backbone) and a classification head, such that
\[
f(\boldsymbol{x}) = \text{Head} \circ \text{Backbone}(\boldsymbol{x}).
\]
Specifically, as illustrated in Fig.~\ref{fig:pipeline}, StableTTA-II modifies the standard inference pipeline by introducing feature-level \textbf{deterministic} cropping and logit aggregation:
\[
\overline{\boldsymbol{z}} = \frac{1}{N} \sum_{i=1}^{N} \text{Head}(\text{crop}^{(i)} (\text{Backbone}(\boldsymbol{x}))).
\]
As an example, we define the crops on the ResNet feature map \(\boldsymbol{h} = \text{Backbone}(\boldsymbol{x})\) as:
\[
\text{crop}^{(0)} (\boldsymbol{h}) = \boldsymbol{h},\quad \text{crop}^{(1)} (\boldsymbol{h}) = \boldsymbol{h}[k:-k, :],\quad \text{crop}^{(2)} (\boldsymbol{h}) = \boldsymbol{h}[:, k:-k].
\]
where \(k\) corresponds to approximately 6.25\% ( = \(1/2 \times (256 - 224)/256\)) of the feature map size along each spatial dimension. This choice is consistent with common practice in ImageNet-trained models, where an 87.5\% ( = 1 - 2\(\times\)6.25\%) crop ratio is typically adopted during both training (random cropping) and evaluation (center cropping).

In Section \ref{sec:Evaluations of StableTTA-II}, we present a detailed analysis of model performance gains using StableTTA-II. The theoretical analysis of computational cost, model head configurations, and feature-level cropping settings are provided in Appendix~\ref{app:StableTTA-II: Implementation Details}. In Appendix~\ref{app:StableTTA-II: Extra Experiments}, we conduct a sensitivity analysis to explore other feature-level augmentations. These experiments show that random cropping and random erasing degrade the final predictions, reducing accuracy below that of the base model.

\section{Experiments}

We evaluate StableTTA in terms of accuracy, memory usage, and computational cost. First, we compare StableTTA across a wide range of ImageNet-1K models. We then show that StableTTA-I enables lightweight architectures to outperform larger models in top-1 accuracy while maintaining favorable efficiency–performance trade-offs. We further present ablation studies and sensitivity analysis to examine the effectiveness and robustness of the proposed methods. Finally, we demonstrate consistent performance improvements achieved by StableTTA-II and provide additional analysis.

\paragraph{Dataset.}
ImageNet-1K \citep{deng2009imagenet} is a commonly used and highly challenging benchmark dataset, which we adopt for evaluation. It contains 1.28 million training images and 50,000 validation images across 1,000 categories. Over the past years, widely accepted models have typically achieved only incremental improvements of 1\%–2\% in top-1 accuracy per advancement \citep{krizhevsky2012imagenet,simonyan2014very,he2016deep,tan2019efficientnet,dosovitskiy2020image}.

\paragraph{Models.}
All pretrained models used in our experiments are publicly available in torchvision \citep{torchvision2016}, including their official implementations, subsequent improvements, and recent training recipes that achieve state-of-the-art performance \citep{torchvision2021sota}. We denote these enhanced variants with ``\textdagger" in this paper.

\paragraph{Computing Platform.} We evaluate model performance on a single NVIDIA RTX A6000 GPU.

\subsection{Evaluations of StableTTA-I}
\label{sec:Evaluations of StableTTA-I}
In our experiments, we set the batch size to 16. Since the ImageNet-1K validation set contains 50 images per class and we use a sequential sampler, the average proportion of samples belonging to the same class in a batch is 92\% (see Appendix~\ref{app:Sequential Sampling and Data Augmentation Policies of StatbleTTA-I}). In Appendix~\ref{app:Sensitivity Analysis of StableTTA-I}, we further discuss the results for ratios of 84\% and 100\%. \textit{Since the sequential sampler can greatly improve model performance, we provide the traditional TTA results obtained with the same sequential sampler in Appendix~\ref{app:Comparison Experiments of StableTTA-I}, Table~\ref{tab:survey_ours}, as a fair comparison to illustrate the advantage of StableTTA-I.}

\begin{figure}[t]
  \centering
  \includegraphics[width=0.975\columnwidth]{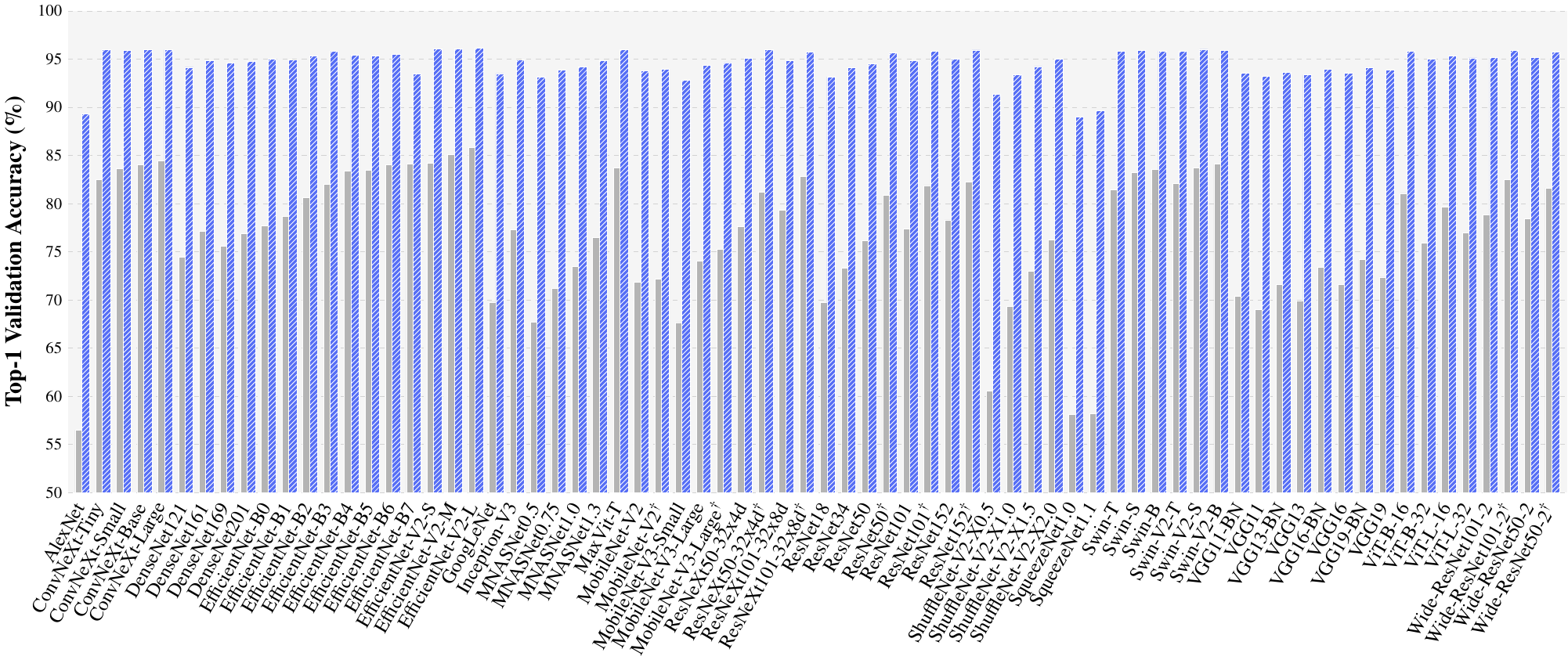}
  \caption{\textbf{Comparison under coherent-batch inference.} StableTTA-I improves model performance across a broad range of ImageNet-1K architectures under coherent-batch inference, highlighting the potential benefits of exploiting deployment-time semantic coherence during test-time aggregation.}
  \label{fig:main_results}
\end{figure}

\begin{figure}
  \begin{center}
    \centerline{
        \begin{tikzpicture}
        \node[inner sep=0pt, outer sep=0pt] (img) {%
          \includegraphics[
            width=0.95\columnwidth,
            trim=2.5cm 31.5cm 1.8cm 2.0cm,
            clip
          ]{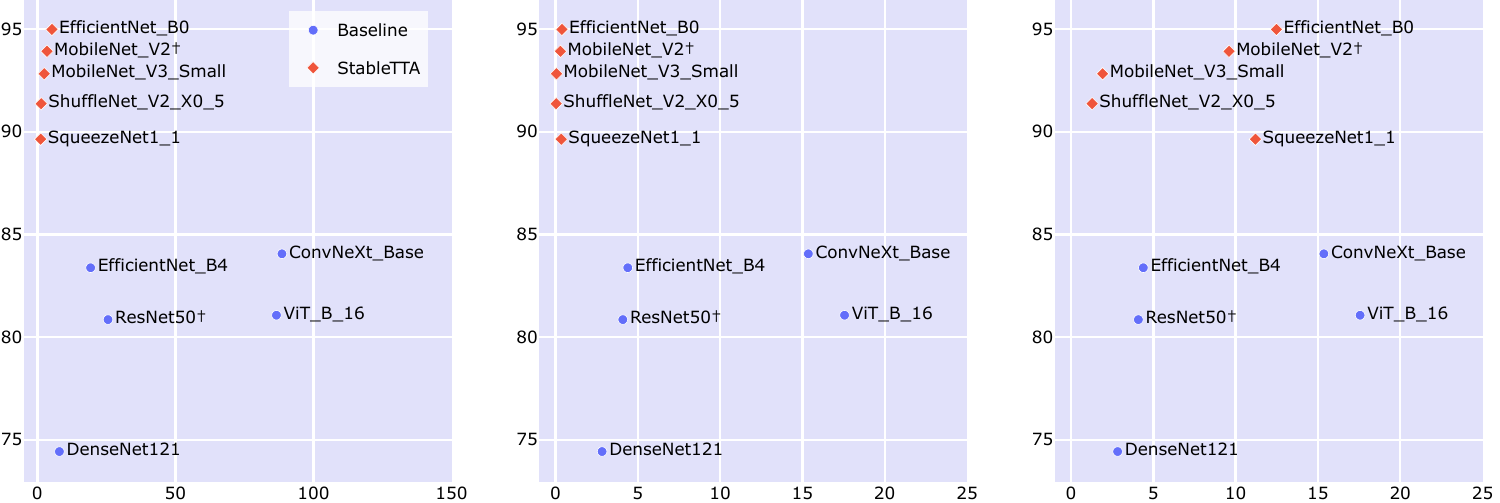}%
        };
        \node[rotate=90, anchor=south, font=\small] at (img.west) {Top-1 Validation Accuracy (\%)};
        \node[anchor=north, xshift=-4.5cm, font=\footnotesize] at (img.south) {Params (Millions)};
        \node[anchor=north, xshift=0.1cm, font=\footnotesize] at (img.south) {Peak GFLOPs (Sequential Mode)};
        \node[anchor=north, xshift=4.625cm, font=\footnotesize] at (img.south) {Total GFLOPs (Parallel Mode)};
        \end{tikzpicture}
    }
    \caption{\textbf{Superior Efficiency and Accuracy of StableTTA-I.} Comparison of the baseline (blue) and StableTTA-I (red) across the number of model parameters (left), peak GFLOPs in sequential aggregation mode (middle), and total GFLOPs in parallel aggregation mode (right).}
    \label{fig:main_advantages}
  \end{center}
\end{figure}

Fig.~\ref{fig:main_results} shows that StableTTA-I consistently improves prediction accuracy across diverse ImageNet-1K models under the coherent-batch inference setting. We also report model parameter counts, computational cost, and detailed values in Appendix~\ref{app:Comparison Experiments of StableTTA-I} (see Table~\ref{tab:main_results}).
Fig.~\ref{fig:main_advantages} shows that these improvements are achieved without increasing model parameters in the sequential setting of multiple forward passes, highlighting the efficiency of StableTTA-I. In the parallel setting, StableTTA-I increases inference cost by a factor of $N$. However, when applied to lightweight architectures, it remains significantly more economical than using large models. This demonstrates a favorable trade-off between accuracy, memory usage, and computational cost. These results suggest that StableTTA-I effectively bridges the gap between lightweight and high-performance models, enabling practical deployment in resource-constrained environments.

\begin{figure}
  \centering
  \begin{subfigure}{0.495\columnwidth}
    \centering
    \includegraphics[width=\columnwidth]{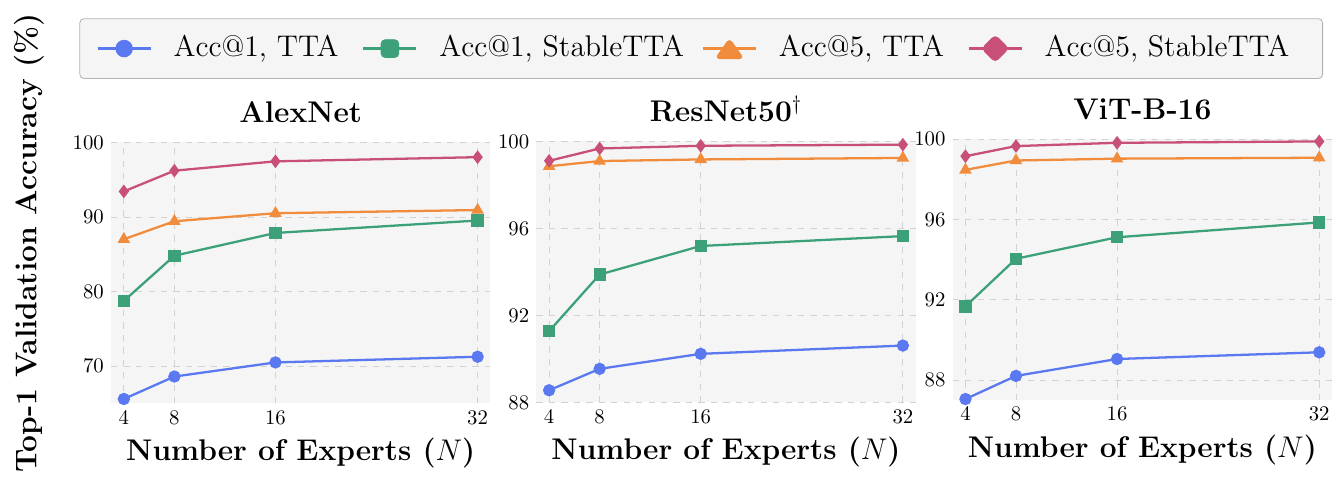}
    \caption{Ablation Study}
    \label{fig:ablation_study}
  \end{subfigure}
  \begin{subfigure}{0.485\columnwidth}
    \centering
    \includegraphics[width=\columnwidth]{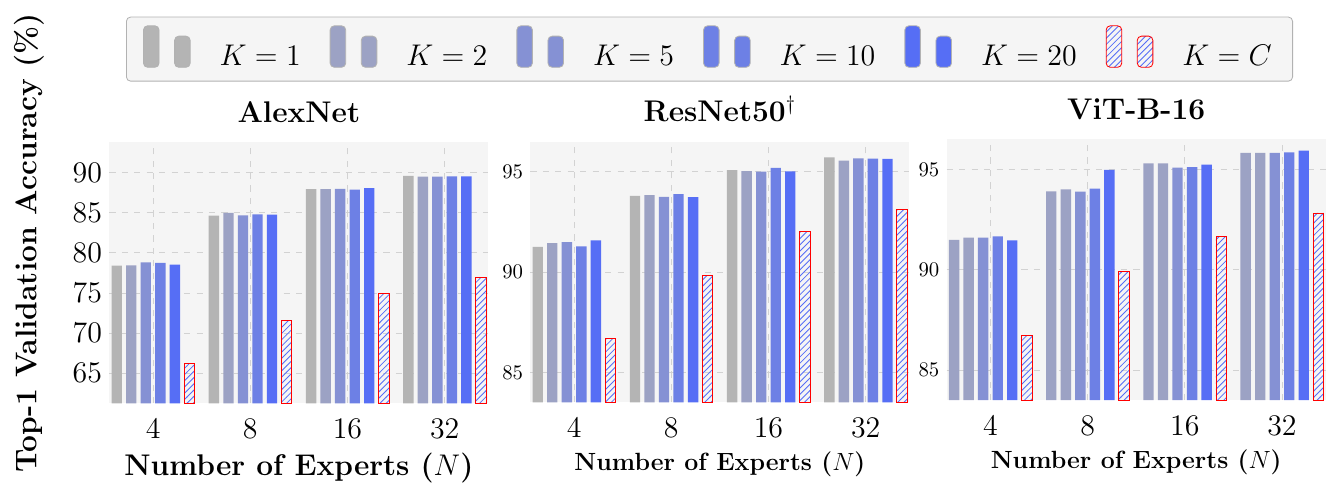}
    \caption{Sensitivity Analysis}
    \label{fig:sensitivity_analysis}
  \end{subfigure}

  \caption{(a) TTA (with our sampler and augmentation) vs. StableTTA-I. (b) StableTTA-I is robust to \(K\), but disabling logit processing \((K=C)\) will significantly reduce accuracy.}
  \label{fig:combined}
\end{figure}

\paragraph{Ablation Study.} We compare standard TTA using sequential sampling with our proposed StableTTA-I. As shown in Fig.~\ref{fig:ablation_study}, StableTTA-I consistently outperforms standard TTA across different models, achieving higher top-1 and top-5 accuracy while maintaining consistent performance gains as \(N\) increases. Detailed ablation results are provided in Appendix~\ref{app:Ablation Study of StableTTA-I}.

We also observed that performance gains saturate as \(N\) increases. So using a large \(N\) yields only marginal improvements. Considering the trade-off with computational cost, we adopt \(N=32\) as a practical setting that balances performance and computational cost.

TTA methods employ flipping, cropping, or image translation as augmentations, resulting in improvements of classification accuracy that are modest (typically less than 2\%) but require 5–100+ times additional computational cost (see Table~\ref{tab:survey_prior} of Appendix~\ref{app:Preliminary Concepts and Reproduction of Prior Work}). 
In Table~\ref{tab:survey_ours} of Appendix~\ref{app:Comparison Experiments of StableTTA-I} and Table~\ref{tab:ablation_study} of Appendix~\ref{app:Ablation Study of StableTTA-I}, we also evaluate conventional TTA methods under the coherent-batch inference setting by adopting the same sequential sampling strategy used in StableTTA-I. Although coherent-batch inference improves the performance of all TTA methods, the gains remain smaller than those achieved by StableTTA-I. These results suggest that both the revised augmentation policy and the proposed logit processing method provide additional benefits beyond coherent-batch aggregation alone.

\paragraph{Sensitivity Analysis.} We analyze the sensitivity of StableTTA-I performance with respect to its key hyperparameters, including the number of candidates \(K\) and the number of experts \(N\). As shown in Fig.~\ref{fig:sensitivity_analysis}, StableTTA-I demonstrates strong robustness to the choice of \(N\), with performance remaining stable across a wide range of values. In contrast, the number of candidates \(K\) has only a marginal effect on top-1 accuracy, indicating that StableTTA-I does not strongly depend on \(K\).

It is important to note that, when all probabilities are retained for logit averaging (\(K = C\)), our NSS is disabled, leading to a significant performance drop and highlighting the critical role in achieving accuracy gains. These results suggest that StableTTA-I provides stable and reliable improvements without requiring meticulous hyperparameter tuning. Additional sensitivity analysis results are provided in Appendix~\ref{app:Sensitivity Analysis of StableTTA-I}.

\subsection{Evaluations of StableTTA-II}
\label{sec:Evaluations of StableTTA-II}

StableTTA-II is a deterministic algorithm that does not rely on sampler stochasticity, batch size, or random data augmentation. It enables improved prediction accuracy for a single image within a single forward pass. Table~\ref{tab:main_results_ii} presents the experimental results compared with the base models, demonstrating improvements across nearly all models.

The advantage of StableTTA-II lies in its simplicity and low computational cost. Using ResNet-50 as an example, its accuracy in traditional TTA mode with 10-crop augmentation reaches 77.15\%, but at the cost of increasing the computational complexity of a single forward pass by 900\% (see Table~\ref{tab:survey_prior} of Appendix \ref{app:Preliminary Concepts and Reproduction of Prior Work}). In contrast, our method improves accuracy to 77.460\%, while introducing only a 0.01\% computational overhead per forward pass.
In Table~\ref{tab:survey_prior} of Appendix \ref{app:Preliminary Concepts and Reproduction of Prior Work}, we also report results from other previously published test-time augmentation methods for further comparison. Notably, these methods incur computational costs that are \textbf{multiple times higher}.

Although conceptually simple, several aspects merit further investigation. For example, our experiments show that replacing the deterministic feature-level cropping described in Section~\ref{sec:StableTTA-II} with other values of \(k\) or random cropping leads to suboptimal results. Similarly, applying random erasing or increasing the number of “experts” degrades performance.

Furthermore, Appendix~\ref{app:StableTTA-II: Implementation Details} provides a detailed description of model head configurations and feature-level cropping settings tailored to different model architectures, along with the formulation of the model head's computation cost. In Appendix~\ref{app:Additional Discussion on Addressing the Computational Cost of Ensemble Methods}, we further discuss possible solutions to reducing the computational cost of ensemble methods.

\begin{longtable}{lr|ll|ll}
\caption{Comparison of base models and the StableTTA-II-enhanced models on the ImageNet-1K.}
\label{tab:main_results_ii} \\
\toprule
& & \multicolumn{2}{c|}{\textbf{Baseline (\%)}} & \multicolumn{2}{c}{\textbf{StableTTA-II (\%)}} \\
\textbf{Weight} & \textbf{Params} & \textbf{Acc@1} & \textbf{Acc@5} & \textbf{Acc@1} & \textbf{Acc@5} \\
\midrule
\endfirsthead

\toprule
\textbf{Weight} & \textbf{Params} & \textbf{Acc@1} & \textbf{Acc@5} & \textbf{Acc@1} & \textbf{Acc@5} \\
\midrule
\endhead

\midrule
\multicolumn{6}{r}{\textit{Continued on next page}} \\
\endfoot

\bottomrule
\endlastfoot


ConvNeXt\_Tiny \cite{liu2022convnet,torchvision2016} & 28.6M & 82.520 & 96.146 & 82.570 & 96.218 \\ \rowcolor{gray!25!white}
ConvNeXt\_Small \cite{liu2022convnet,torchvision2016} & 50.2M & 83.616 & 96.65 & 83.692 & 96.778 \\
ConvNeXt\_Base \cite{liu2022convnet,torchvision2016} & 88.6M & 84.062 & 96.87 & 84.208 & 96.952 \\ \rowcolor{gray!25!white}
ConvNeXt\_Large \cite{liu2022convnet,torchvision2016} & 197.8M & 84.414 & 96.976 & 84.524 & 97.128 \\

DenseNet121 \cite{huang2017densely,torchvision2016} & 8.0M & 74.434 & 91.972 & 75.980 & 93.062 \\ \rowcolor{gray!25!white}
DenseNet161 \cite{huang2017densely,torchvision2016} & 28.7M & 77.138 & 93.56 & 78.586 & 94.366 \\ 
DenseNet169 \cite{huang2017densely,torchvision2016} & 14.1M & 75.6 & 92.806 & 77.128 & 93.662 \\ \rowcolor{gray!25!white}
DenseNet201 \cite{huang2017densely,torchvision2016} & 20.0M & 76.896 & 93.37 & 78.128 & 94.146 \\

EfficientNet\_B0 \cite{tan2019efficientnet,torchvision2016} & 5.3M & 77.692 & 93.532 & 79.074 & 94.408 \\ \rowcolor{gray!25!white}
EfficientNet\_B1 \cite{tan2019efficientnet,torchvision2016} & 7.8M & 78.642 & 94.186 & 79.446 & 94.634 \\
EfficientNet\_B2 \cite{tan2019efficientnet,torchvision2016} & 9.1M & 80.608 & 95.31 & 81.014 & 95.454 \\ \rowcolor{gray!25!white}
EfficientNet\_B3 \cite{tan2019efficientnet,torchvision2016} & 12.2M & 82.008 & 96.054 & 82.322 & 96.128 \\
EfficientNet\_B4 \cite{tan2019efficientnet,torchvision2016} & 19.3M & 83.384 & 96.594 & 83.060 & 96.306 \\ \rowcolor{gray!25!white}

MNASNet0\_5 \cite{tan2019mnasnet,torchvision2016} & 2.2M & 67.734 & 87.49 & 69.624 & 88.942 \\
MNASNet0\_75 \cite{tan2019mnasnet,torchvision2016} & 3.2M & 71.18 & 90.496 & 71.482 & 90.818 \\ \rowcolor{gray!25!white}
MNASNet1\_0 \cite{tan2019mnasnet,torchvision2016} & 4.4M & 73.456 & 91.51 & 75.056 & 92.364 \\
MNASNet1\_3 \cite{tan2019mnasnet,torchvision2016} & 6.3M & 76.506 & 93.522 & 76.720 & 93.736 \\ \rowcolor{gray!25!white}

MobileNet\_V2 \cite{sandler2018mobilenetv2,torchvision2016} & 3.5M & 71.878 & 90.286 & 73.582 & 91.594 \\
MobileNet\_V3\_Small \cite{howard2019searching,torchvision2016} & 2.5M & 67.668 & 87.402 & 69.570 & 88.774 \\ \rowcolor{gray!25!white}
MobileNet\_V3\_Large \cite{howard2019searching,torchvision2016} & 5.5M & 74.042 & 91.34 & 75.430 & 92.218 \\

ResNeXt50\_32x4d \cite{xie2017aggregated,torchvision2016} & 25.0M & 77.618 & 93.698 & 79.108 & 94.500 \\ \rowcolor{gray!25!white}
ResNeXt101\_32x8d \cite{xie2017aggregated,torchvision2016} & 88.8M & 79.312 & 94.526 & 80.478 & 95.278 \\
ResNeXt101\_64x4d \cite{xie2017aggregated,torchvision2016} & 88.8M & 83.246 & 96.454 & 83.110 & 96.558 \\ \rowcolor{gray!25!white}

ResNet18 \cite{he2016deep,torchvision2016} & 11.7M & 69.758 & 89.078 & 71.540 & 90.364 \\
ResNet34 \cite{he2016deep,torchvision2016} & 21.8M & 73.314 & 91.42 & 75.060 & 92.500 \\ \rowcolor{gray!25!white}
ResNet50 \cite{he2016deep,torchvision2016} & 25.6M & 76.130 & 92.862 & 77.460 & 93.738 \\
ResNet101 \cite{he2016deep,torchvision2016} & 44.5M & 77.374 & 93.546 & 78.636 & 94.534 \\ \rowcolor{gray!25!white}
ResNet152 \cite{he2016deep,torchvision2016} & 60.2M & 78.312 & 94.046 & 79.668 & 94.892 \\

ShuffleNet\_V2\_X0\_5 \cite{ma2018shufflenet,torchvision2016} & 1.4M & 60.552 & 81.746 & 62.292 & 83.200 \\ \rowcolor{gray!25!white}
ShuffleNet\_V2\_X1\_0 \cite{ma2018shufflenet,torchvision2016} & 2.3M & 69.362 & 88.316 & 70.974 & 89.570 \\
ShuffleNet\_V2\_X1\_5 \cite{ma2018shufflenet,torchvision2016} & 3.5M & 72.996 & 91.086 & 73.570 & 91.638 \\ \rowcolor{gray!25!white}
ShuffleNet\_V2\_X2\_0 \cite{ma2018shufflenet,torchvision2016} & 7.4M & 76.23 & 93.006 & 76.862 & 93.492 \\

Wide\_ResNet50\_2 \cite{zagoruyko2016wide,torchvision2016} & 68.9M & 78.468 & 94.086 & 79.794 & 94.780 \\ \rowcolor{gray!25!white}
Wide\_ResNet101\_2 \cite{zagoruyko2016wide,torchvision2016} & 126.9M & 78.848 & 94.284 & 80.316 & 95.106 \\


\end{longtable}

\section{Limitations and Future Work}

StableTTA-I assumes coherent-batch inference, where neighboring observations are semantically correlated. While this setting naturally arises in applications such as burst imaging, robotics, industrial inspection, and video perception, it does not apply to fully IID inference scenarios. Consequently, the applicability of StableTTA-I depends on deployment conditions that exhibit local semantic consistency.
In addition, StableTTA-I still relies on multiple forward passes for aggregation and therefore inherits the computational overhead of conventional TTA methods. Although the method remains substantially more efficient than large-model ensembles and can be combined with lightweight architectures, reducing inference cost remains an important direction for future work.
Compared with StableTTA-I, StableTTA-II provides smaller performance gains but imposes almost no practical deployment constraints. It operates with a single forward pass and negligible additional computation, although its current formulation assumes CNN-style intermediate feature tensors.
For future work, we plan to extend StableTTA to other tasks, such as semantic segmentation \citep{long2015fully} and object detection \citep{ren2015faster}, and explore its integration with techniques such as model pruning \citep{han2015learning} and knowledge distillation \citep{hinton2015distilling}.

\section{Conclusion}
\label{sec:Conclusion}

In this work, we investigate aggregation inconsistency in ensemble-based inference and show that nonlinear projection through softmax and voting operations can produce unstable predictions across aggregation strategies. We further demonstrate that this inconsistency is closely related to logit variance under coherent-batch inference settings.
Motivated by this observation, we propose StableTTA-I, a training-free test-time adaptation method that improves aggregation stability through variance-aware logit processing and coherent-batch aggregation. Across diverse ImageNet-1K models, StableTTA-I produces substantial accuracy gains while maintaining favorable efficiency–performance trade-offs relative to large-model ensembles.
We additionally introduce StableTTA-II, a lightweight single-forward-pass aggregation framework using feature-level cropping, enabling prediction enhancement with negligible computational overhead.
Overall, our results suggest that aggregation stability provides a useful perspective for understanding and improving test-time adaptation and ensemble inference in practical visual recognition systems.

\bibliographystyle{plainnat}
\bibliography{references}


\appendix

\section{Preliminary Concepts and Reproduction of Prior Work}
\label{app:Preliminary Concepts and Reproduction of Prior Work}

In this section, we review key concepts and reproduce prior TTA results to highlight their limitations and motivate our method.

\begin{figure}[ht]
    \centering 
    \includegraphics[width=\columnwidth]{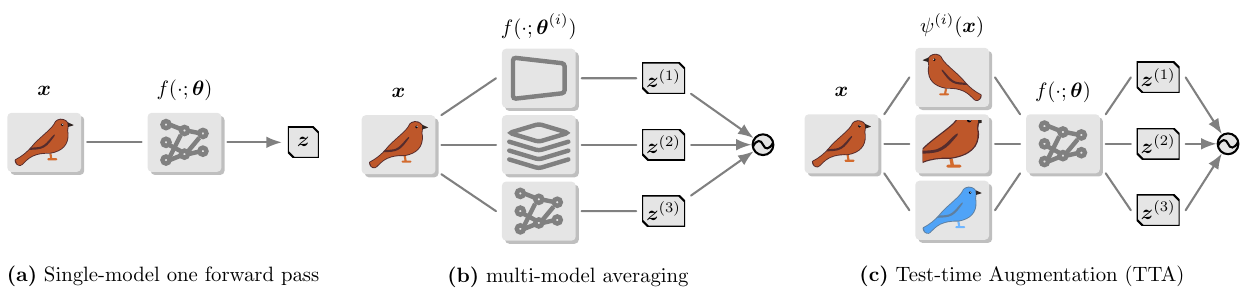} 
    \caption{
    (a) \textbf{Baseline:} a single model with one forward pass.
    (b) \textbf{Multi-model ensemble:} multiple models with multiple forward passes.
    (c) \textbf{TTA:} a single model with multiple forward passes using augmented inputs.
    Both the multi-model ensemble and TTA aggregate outputs across forward passes.} 
    \label{fig:background}
\end{figure}

Fig.~\ref{fig:background} illustrates three common inference strategies for model predictions. (a) shows the baseline approach, where a single model processes the input once to produce an output. (b) presents a multi-model ensemble, in which predictions from several independently trained models are combined to improve accuracy, at the cost of increasing the total model size and the computational overhead. (c) demonstrates TTA, where multiple augmented images of the same input are passed through a single model and their predictions are aggregated. While both ensemble methods (b) and (c) can enhance performance compared to the baseline (a), they require multiple forward passes, leading to higher inference cost.

In addition, Table~\ref{tab:survey_prior} summarizes representative TTA results reported on the ImageNet-1K validation set. A range of augmentation policies, such as AutoAugment, flipping, cropping, and GPS, have been evaluated across diverse architectures. Although these methods consistently improve accuracy over baseline (single-pass inference), the gains remain limited: all reported results stay below 80\% top-1 accuracy, incurring increased inference cost due to multiple forward passes.

\begin{table}[ht]
\caption{Summary of experimental results reported in prior studies, evaluated using top-1 accuracy on the ImageNet-1K validation set. Both baseline and test-time augmentation (TTA) results are taken directly from the original papers, with each value corresponding to its respective reference.}
\label{tab:survey_prior} 
\centering
\begin{tabular}{l|rr|ll}
\toprule
\textbf{Weight} & \textbf{Data Augmentation} & \textbf{Baseline} & \textbf{TTA Acc} & \textbf{Source paper} \\
\midrule

VGG(v5) & 10 \(\times \) Crop & & 71.93 & \citep{he2016deep} He et al. \\
PReLU-net & 10 \(\times \) Crop & & 75.73 & \citep{he2016deep} He et al. \\
plain-34 & 10 \(\times \) Crop & & 71.46 & \citep{he2016deep} He et al. \\
ResNet50 & 10 \(\times \) Crop & & 77.15 & \citep{he2016deep} He et al. \\
ResNet101 & 10 \(\times \) Crop & & 78.25 & \citep{he2016deep} He et al. \\
ResNet152 & 10 \(\times \) Crop & & 78.57 & \citep{he2016deep} He et al. \\
DenseNet-121 & 10 \(\times \) Crop & 74.98 & 76.39 & \citep{huang2017densely} Huang et al. \\
DenseNet-169 & 10 \(\times \) Crop & 76.20 & 77.92 & \citep{huang2017densely} Huang et al. \\
DenseNet-201 & 10 \(\times \) Crop & 77.42 & 78.54 & \citep{huang2017densely} Huang et al. \\
DenseNet-264 & 10 \(\times \) Crop & 77.85 & 79.20 & \citep{huang2017densely} Huang et al. \\
MobileNetV2 & 30 \(\times \) AutoAugment & & 71.38 & \citep{shanmugam2021better} Shanmugam et al. \\ 
InceptionV3 & 30 \(\times \) AutoAugment & & 69.51 & \citep{shanmugam2021better} Shanmugam et al. \\  
ResNet18 & 30 \(\times \) AutoAugment & & 69.62 & \citep{shanmugam2021better} Shanmugam et al. \\
ResNet50 & 30 \(\times \) AutoAugment & & 75.53 & \citep{shanmugam2021better} Shanmugam et al. \\
MobileNetV2 & 128 \(\times \) (AutoAugment+\citep{shanmugam2021better}) & & 72.57 & \citep{shanmugam2021better} Shanmugam et al.  \\ 
InceptionV3 & 128 \(\times \) (AutoAugment+\citep{shanmugam2021better}) & & 71.02 & \citep{shanmugam2021better} Shanmugam et al. \\  
ResNet18 & 128 \(\times \) (AutoAugment+\citep{shanmugam2021better}) & & 70.89 & \citep{shanmugam2021better} Shanmugam et al. \\
ResNet50 & 128 \(\times \) (AutoAugment+\citep{shanmugam2021better}) & & 76.36 & \citep{shanmugam2021better} Shanmugam et al. \\
ResNet50 & 5 \(\times\) Crop & & 76.19 & \citep{kim2020learning} Kim et al. \\
ResNet50 & 10 \(\times\) Crop & & 76.94 & \citep{kim2020learning} Kim et al. \\
ResNet50 & 4 \(\times\) GPS\citep{lyzhov2020greedy} & & 76.56 & \citep{kim2020learning} Kim et al. \\
ResNet50 & 2 \(\times\) Flip + 2 \( \times \) \citep{kim2020learning} & & 77.90 & \citep{kim2020learning} Kim et al. \\
ResNet50 & 5 \(\times\) Crop + 2 \(\times\) \citep{kim2020learning} & & 79.34 & \citep{kim2020learning} Kim et al. \\

\bottomrule
\end{tabular}
\end{table}

\section{Sequential Sampling and Data Augmentation Policies of StatbleTTA-I}
\label{app:Sequential Sampling and Data Augmentation Policies of StatbleTTA-I}

\paragraph{Sequential Sampling.} The ImageNet-1K validation set contains 50,000 images, organized into 1,000 classes with 50 images per class, stored sequentially by class. When sampling this dataset sequentially using a small batch size (without shuffling), batches are not class-balanced: most batches contain samples predominantly from a single class, with occasional boundary batches that mix two adjacent classes. 
For example, when the batch size is set to 16, the dataset is partitioned into groups of image-label pairs as follows:
\[
\underbrace{(\text{img}_1, C_1), \dots}_{\text{1st batch}},
\underbrace{(\text{img}_{17}, C_1), \dots}_{\text{2nd batch}},
\underbrace{(\text{img}_{33}, C_1), \dots}_{\text{3rd batch}},
\underbrace{(\text{img}_{49}, C_1), \dots, (\text{img}_{64}, C_2)}_{\text{4th batch}}, \dots
\]
Since batch boundaries and class boundaries repeat every
\[
\operatorname{lcm}(50, 16) = 400,
\]
each cycle contains \(400/16=25\) batches. Within each cycle, the number of samples from the majority class in each batch follows the pattern:
\[
16, 16, 16, 14, 16, 16, 12, 16, 16, 10, 16, 16, 8, 16, 16, 10, 16, 16, 12, 16, 16, 14, 16, 16, 16.
\]
The average majority-class ratio is \(368/400=92\%\).

\paragraph{Augmentation Policies.} Since neural network layers are continuous and differentiable, most neural networks satisfy the Hölder condition~\citep{virmaux2018lipschitz}:
\[
    \| \boldsymbol{z} - \boldsymbol{z}' \| = \| f(\boldsymbol{x};\boldsymbol{\theta}) - f(\boldsymbol{x}';\boldsymbol{\theta}) \| \le c \cdot \| \boldsymbol{x} - \boldsymbol{x}' \|^d,
\]
where \(c\) and \(d\) are some constants. This inequality indicates that the distance \(\| \boldsymbol{z} - \boldsymbol{z}' \|\) between logits is bounded by the distance \(\| \boldsymbol{x} - \boldsymbol{x}' \|\) between inputs.

Now, consider a sequence of data augmentations \(\{ \psi_1, \psi_2, \dots \}\) and the corresponding logit vector \(\boldsymbol{z}^{(i)} := f(\psi_i(\boldsymbol{x}); \boldsymbol{\theta})\). By applying the Hölder condition, we obtain:
\begin{equation}
    \| \boldsymbol{z}^{(i)} - \boldsymbol{z}^{(j)} \| =
    \| f(\psi_i(\boldsymbol{x});\boldsymbol{\theta}) - f(\psi_j(\boldsymbol{x});\boldsymbol{\theta}) \| \le 
    c \cdot \| \psi_i(\boldsymbol{x}) - \psi_j(\boldsymbol{x}) \|^d.
    \label{eq:holder_condition}
\end{equation}
To facilitate understanding Eq.~(\ref{eq:holder_condition}), in Appendix \ref{app:Hölder Condition of StableTTA-I}, we provide Fig. \ref{fig:holder_condition} for visualization, along with empirical results as supporting evidence.

When \(d \in (0,1]\), since \(\boldsymbol{z},\boldsymbol{z}'\) are independent and identically distributed, and using the identity \(\mathbb{E}\|\boldsymbol{z} - \boldsymbol{z}'\|^2 = 2\,\mathrm{Var}(\boldsymbol{z})\), Jensen’s inequality gives: 
\[
\text{Var}(\boldsymbol{z}) \le 2^{d-1}c^2 \text{Var}(\psi(\boldsymbol{x})).
\]

Now, revisiting the inconsistency discussed in Section~\ref{sec:Background}, we evaluate data augmentation policies as follows:
(1) Simple augmentation methods such as horizontal flipping and random cropping are not necessary in TTA, as the model is not sensitive to them (see Table \ref{tab:survey_ours} of Appendix \ref{app:Comparison Experiments of StableTTA-I}). These techniques are standard in model training and provide only marginal improvements. Consequently, the resulting logit vectors are typically identical after such augmentations.
(2) Strong augmentation techniques such as translation, rotation, and affine transformations \citep{cubuk2019autoaugment} should be avoided in TTA, as they can induce an excessively large increase in \(\operatorname{Var}  (\psi(\boldsymbol{x})) \).
(3) Although random erasing yields relatively small values of \(\operatorname{Var} ( \psi(\boldsymbol{x}) )\), it provides limited diversity among augmented samples and is therefore excluded.
(4) Mixup \citep{zhang2018mixup} and CutMix \citep{yun2019cutmix} are strong data augmentation methods; however, they operate across image patches, which may alter the ground-truth labels. Therefore, when applying mixup or CutMix, we need to ensure that the batch of images largely belongs to the same class.

Under these considerations, mixup and CutMix emerge as promising candidates. However, both require further modification to reduce \(\operatorname{Var} ( \psi(\boldsymbol{x}) )\):

\begin{itemize}
\item \textbf{Mixup:} Instead of combining the input with a random image, we perform a weighted combination with a fixed image, which is randomly sampled and kept unchanged thereafter.
\item \textbf{CutMix:} The covering region is typically sampled from a Beta distribution to encourage diversity during training. In contrast, we fix the window size to one-quarter of the image to reduce \(\operatorname{Var} ( \psi(\boldsymbol{x}) )\).
\item \textbf{Random Choice:} Standard pipelines often compose multiple augmentations. Here, we randomly choose mixup or CutMix in each forward pass to better control \(\operatorname{Var} ( \psi(\boldsymbol{x}) )\).
\end{itemize}

\section{Jarque–Bera Test of StableTTA-I}
\label{app:Jarque–Bera Test of StableTTA-I}

To support the assumption we used in Section~\ref{sec:StableTTA-I}, we introduce the Jarque–Bera test \citep{jarque1980efficient} and then empirically show that the logit \(\boldsymbol{z}_k \sim \mathcal{N}(\boldsymbol{\mu}_k, \sigma_k^2)\), based on our proposed data augmentation policies for any \(k \in \{1, \dots, C\}\). 
We select the Jarque–Bera test, which is suitable for small sample sizes, with \(N \le 32\) in our setting.

For a given \(k \in \{1, \dots, C\}\), we consider the hypothesis:
\[
H_0: \boldsymbol{z}_k^{(1)}, \dots, \boldsymbol{z}_k^{(N)} \overset{\text{i.i.d.}}{\sim} \mathcal{N}(\mu_k, \sigma_k^2),
\]
where \(\mu_k \in \mathbb{R}\) and \(\sigma_k > 0\) are unknown parameters. In Jarque–Bera test, under \(H_0\):
\[
JB = \frac{N}{6} \left( S^2 + \frac{(K-3)^2}{4} \right) \overset{\text{approx.}}{\sim} \mathcal{X}^2(2),
\]
where \(S\) is the skewness and \(K\) is the kurtosis. Therefore, at significance level \(\alpha=0.05\), we reject \(H_0\) if:
\[
JB > \mathcal{X}^2_{0.95, 2} \approx 5.99.
\]

\begin{figure}[!t]
  \centering
  \includegraphics[width=\columnwidth]{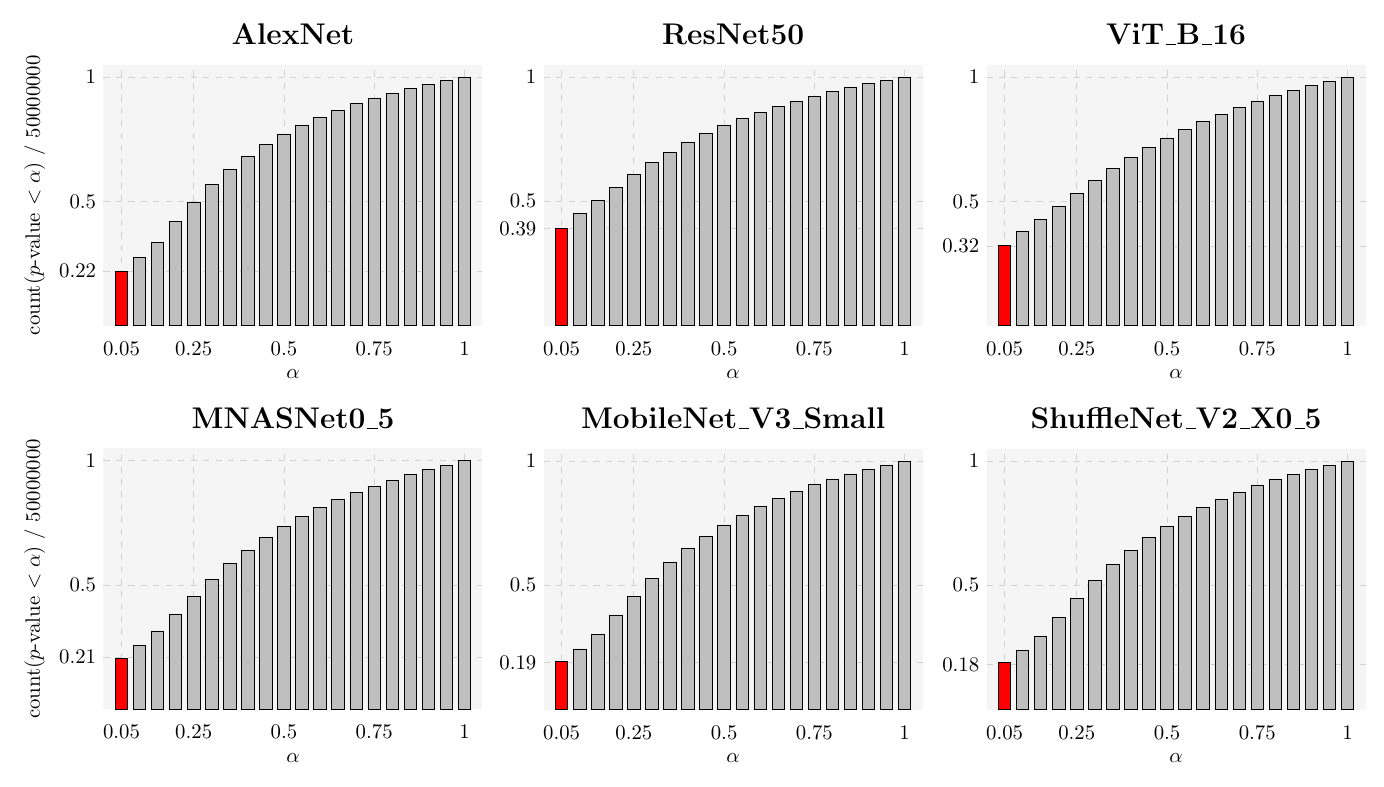}
  \caption{Empirical cumulative distribution functions (ECDFs) of \(p\)-values from Jarque–Bera tests across models on ImageNet-1K validation set. 
  Each model outputs 50,000\(\times\)1,000 groups from augmented images, 
  with one test per group. The x-axis shows the significance level \(\alpha\), and the y-axis shows the proportion of \(p\)-values \(\le \alpha\). The red bar marks the case of \(\alpha=0.05\). 
  A lower proportion (e.g., 22\% of the tested groups under AlexNet) indicates that the majority of groups fail to reject the null hypothesis, suggesting stronger consistency with a normal distribution.
  }
  \label{fig:statistic}
\end{figure}

Based on this, we design a Monte Carlo test to verify the hypothesis. Since the ImageNet-1K validation set contains 50,000 images across 1,000 classes, we obtain 50,000\(\times\)1,000 groups, each of which consists of \(N\) individual data points \(\{\boldsymbol{z}_k^{(1)}, \dots, \boldsymbol{z}_k^{(N)}\}\). For each group, a Jarque–Bera test is conducted to evaluate the \(H_0\) that the samples follow a normal distribution. To better understanding, we perform \(N\) individual data augmentations on the 50,000 images. The AlexNet model then outputs \(N\times\)50,000 logit vectors with the dimension of 1,000. Thus, we obtain 50,000\(\times\)1,000 groups (each with \(N\) samples), and our goal is to test whether each group follows a normal distribution. Consequently, for each model shown in Fig.~\ref{fig:statistic}, we plot the empirical cumulative distribution function (cumulative histogram) of the resulting 50,000,000 \(p\)-values, where the x-axis represents significance levels (\(\alpha\)) ranging from 0 to 1, and the y-axis shows the cumulative proportion of tests with \(p\)-values less than or equal to each \(\alpha\). These histograms present an overall assessment of how often the normality assumption holds across all images and classes. 

The figure shows that, in AlexNet's outputs, 22\% of the groups have \(p\)-values less than 0.05. This implies that 78\% of the groups fail to reject the null hypothesis \(H_0\), suggesting that these groups are consistent with a normal distribution. Similar trends are observed for other models.

\section{Monte Carlo Simulation of StableTTA-I}
\label{app:Monte Carlo Simulation of StableTTA-I}

In this section, we present a Monte Carlo simulation with mathematical derivations and experimental visualizations to support our explanation of the aggregation inconsistency phenomenon introduced in Section~\ref{sec:Background}.

Based on our discussion in Appendix~\ref{app:Jarque–Bera Test of StableTTA-I}, we assume that the logit vector $\boldsymbol{z} \sim \mathcal{N}(\boldsymbol{\mu},\sigma^2 \boldsymbol{I})$ in a binary classification task. Then,
\begin{align}
    \mathbb{E} \begin{pmatrix} \boldsymbol{z}_1 - \boldsymbol{z}_2 \\ \mathds{1}_{\boldsymbol{z}_1 > \boldsymbol{z}_2} \end{pmatrix}
    &= \begin{bmatrix} \boldsymbol{\mu}_1 - \boldsymbol{\mu}_2 \\ \Phi \left( \frac{\boldsymbol{\mu}_1 - \boldsymbol{\mu}_2}{\sqrt{2}\sigma} \right) \end{bmatrix}, \\
    \operatorname{Var} \begin{pmatrix} \boldsymbol{z}_1 - \boldsymbol{z}_2 \\ \mathds{1}_{\boldsymbol{z}_1 > \boldsymbol{z}_2} \end{pmatrix}
    &= \begin{bmatrix} 2\sigma^2 & \sqrt{2}\sigma \phi(a) \\ \sqrt{2}\sigma \phi(a) & \Phi(a)-\Phi^2(a) \end{bmatrix},
\end{align}
where
\[
a = \frac{\boldsymbol{\mu}_1 - \boldsymbol{\mu}_2}{\sqrt{2}\sigma}.
\]
By the Central Limit Theorem, for large $N$, the sample averages satisfy
\[
\begin{pmatrix} \frac{1}{N} \sum\limits_{i=1}^{N} \boldsymbol{z}_1^{(i)} - \boldsymbol{z}_2^{(i)} \\ \frac{1}{N} \sum\limits_{i=1}^{N} \mathds{1}_{\boldsymbol{z}_1^{(i)} > \boldsymbol{z}_2^{(i)}} \end{pmatrix} \sim \mathcal{N} \left( \begin{bmatrix} \boldsymbol{\mu}_1 - \boldsymbol{\mu}_2 \\ \Phi \left( \frac{\boldsymbol{\mu}_1 - \boldsymbol{\mu}_2}{\sqrt{2}\sigma} \right) \end{bmatrix}, \frac{1}{N} \begin{bmatrix} 2\sigma^2 & \sqrt{2}\sigma \phi(a) \\ \sqrt{2}\sigma \phi(a) & \phi(a)-\phi^2(a) \end{bmatrix} \right).
\]
Define the standardized variables
\[
\begin{bmatrix} X \\ Y \end{bmatrix} := \begin{bmatrix} \frac{\frac{1}{N} \sum\limits_{i=1}^{N} (\boldsymbol{z}_1^{(i)} - \boldsymbol{z}_2^{(i)}) - (\boldsymbol{\mu}_1 - \boldsymbol{\mu}_2)}{\sqrt{2}\sigma/\sqrt{N}} \\ \frac{\frac{1}{N} \sum\limits_{i=1}^{N} \mathds{1}_{\boldsymbol{z}_1^{(i)} > \boldsymbol{z}_2^{(i)}} - \Phi(a)}{\sqrt{\Phi(a)-\Phi^2(a)}/\sqrt{N}} \end{bmatrix},
\]
which jointly follow
\[
\begin{bmatrix} X \\ Y \end{bmatrix} \sim \mathcal{N} \left( \begin{bmatrix} 0 \\ 0 \end{bmatrix}, \begin{bmatrix} 1 & \frac{\phi(a)}{\sqrt{\Phi(a)(1-\Phi(a))}} \\ \frac{\phi(a)}{\sqrt{\Phi(a)(1-\Phi(a))}} & 1 \end{bmatrix} \right).
\]

\begin{figure}
  \vskip 0.2in
  \begin{center}
    \centerline{
        \begin{tikzpicture}
        \node[inner sep=0pt, outer sep=0pt] (img) {%
          \includegraphics[
            width=0.95\columnwidth,
            trim=2.5cm 16.75cm 2.5cm 1.8cm,
            clip
          ]{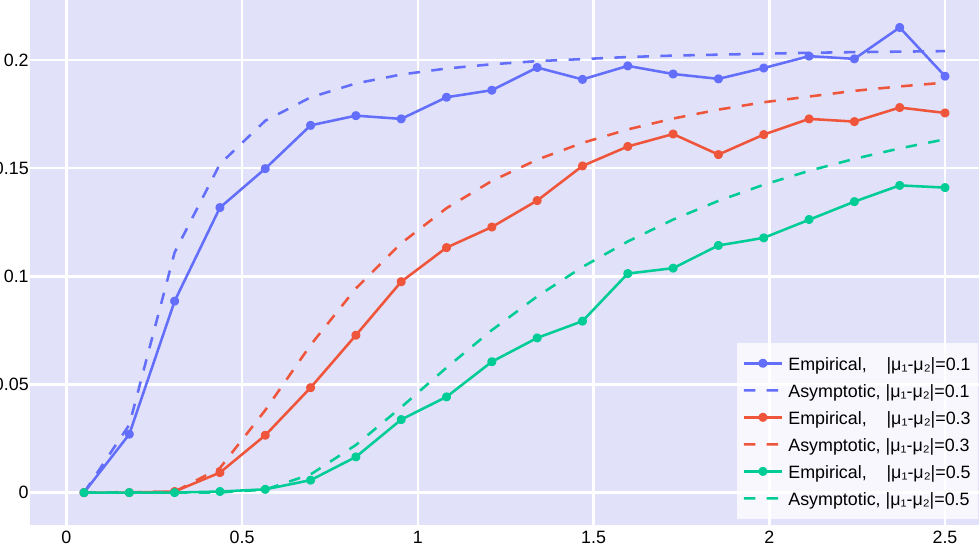}%
        };
        \node[rotate=90, anchor=south] at (img.west) {Inconsistency rate $\operatorname{P}(\hat{y}_{\operatorname{logit}} \neq \hat{y}_{\operatorname{hard}})$};
        \node[anchor=north] at (img.south) {Gaussian noise $\sigma = \sqrt{\operatorname{Var}(\boldsymbol{z})}$};
        \end{tikzpicture}
    }
    \caption{\textbf{Monte Carlo simulation.} The inconsistency probability increases as \(\operatorname{Var}(\boldsymbol{z})\) grows. In this simulation, we consider distributions: \(\{ \boldsymbol{z} \sim \mathcal{N} (\boldsymbol{\mu}, \sigma I) \mid \mu \in \{(1, 0.9), (1, 0.7), (1, 0.5)\}, \sigma \in [0.05, 0.25] \}\). The solid curves show Monte Carlo estimates of the relationship between $\sigma$ and $\operatorname{P}(\hat{y}_{\text{logit}} \neq \hat{y}_{\text{hard}})$, while the dashed curves correspond to the theoretical (asymptotic) predictions. The empirical and theoretical results are closely matched.}
    \label{fig:simulation}
  \end{center}
\end{figure}

We now compute the probability of aggregation inconsistency between the logit-averaging and the hard voting method. First,
\begin{align}
    \operatorname{P}(\hat{y}_{\text{logit}}=1, \hat{y}_{\text{hard}}=2) 
    &= \operatorname{P} \left( \frac{1}{N}\sum_{k=1}^N (\boldsymbol{z}_1^{(k)} - \boldsymbol{z}_2^{(k)}) > 0,\frac{1}{N}\sum_{i=1}^N \mathds{1}_{\boldsymbol{z}_1^{(k)} > \boldsymbol{z}_2^{(k)}} \le \frac{1}{2} \right) \notag \\
    &= P \left( X > -\sqrt{N}a, Y \le \frac{(1/2 - \phi(a))\sqrt{N}}{\sqrt{\Phi(a)-\Phi^2(a)}} \right) \notag \\
    &= \Phi \left( \frac{(1/2 - \Phi(a))\sqrt{N}}{\sqrt{\Phi(a)-\Phi^2(a)}} \right)  \notag \\
    &- \Phi_2 \left( -a\sqrt{N}, \frac{(1/2 - \Phi(a))\sqrt{N}}{\sqrt{\Phi(a)-\Phi^2(a)}}; \frac{\phi(a)}{\sqrt{\Phi(a)-\Phi^2(a)}} \right),
\end{align}
where \(\Phi\) and \(\Phi_2\) denote the cumulative distribution function (CDF) of the standard normal distribution and the joint CDF of the standard bivariate normal distribution, respectively.

Similarly,
\begin{align}
    \operatorname{P}(\hat{y}_{\text{logit}}=2, \hat{y}_{\text{hard}}=1) 
    &= \operatorname{P} \left( \frac{1}{N}\sum_{k=1}^N (\boldsymbol{z}_1^{(k)} - \boldsymbol{z}_2^{(k)}) \le 0,\frac{1}{N}\sum_{i=1}^N \mathds{1}_{\boldsymbol{z}_1^{(k)} > \boldsymbol{z}_2^{(k)}} > \frac{1}{2} \right) \notag \\
    &= P \left( X \le -\sqrt{N}a, Y > \frac{(1/2 - \phi(a))\sqrt{N}}{\sqrt{\Phi(a)-\Phi^2(a)}} \right) \notag \\
    &= \Phi \left( -\sqrt{N}a \right) \notag \\
    &- \Phi_2 \left( -a\sqrt{N}, \frac{(1/2 - \Phi(a))\sqrt{N}}{\sqrt{\Phi(a)-\Phi^2(a)}}; \frac{\phi(a)}{\sqrt{\Phi(a)-\Phi^2(a)}} \right).
\end{align}

Therefore, the total inconsistency probability is
\begin{align}
\operatorname{P}(\hat{y}_{\text{logit}} \neq \hat{y}_{\text{hard}}) 
&= \operatorname{P}(\hat{y}_{\text{logit}}=1, \hat{y}_{\text{hard}}=2) + \operatorname{P}(\hat{y}_{\text{logit}}=2, \hat{y}_{\text{hard}}=1) \notag \\
&= \Phi \left( -\sqrt{N}a \right) + \Phi \left( \frac{(1/2 - \Phi(a))\sqrt{N}}{\sqrt{\Phi(a)-\Phi^2(a)}} \right) \notag \\
&- 2 \Phi_2 \left( -a\sqrt{N}, \frac{(1/2 - \Phi(a))\sqrt{N}}{\sqrt{\Phi(a)-\Phi^2(a)}}; \frac{\phi(a)}{\sqrt{\Phi(a)-\Phi^2(a)}} \right).
\end{align}
This expression indicates that the inconsistency rate increases as $\operatorname{Var}(\boldsymbol{z}) = \sigma^2$ grows. Fig. \ref{fig:simulation} provides empirical support for this conclusion. The solid curves represent Monte Carlo estimates of the relationship between $\sigma$ and $\operatorname{P}(\hat{y}_{\text{logit}} \neq \hat{y}_{\text{hard}})$, while the dashed curves correspond to the theoretical (asymptotic) predictions.  We observe that the empirical and theoretical curves closely match.

\section{Hölder Condition of StableTTA-I}
\label{app:Hölder Condition of StableTTA-I}

In this section, we present a visualization to explain the Hölder condition introduced in Eq. (\ref{eq:holder_condition}).

\begin{figure}[ht]
  \centering
  \includegraphics[width=\columnwidth]{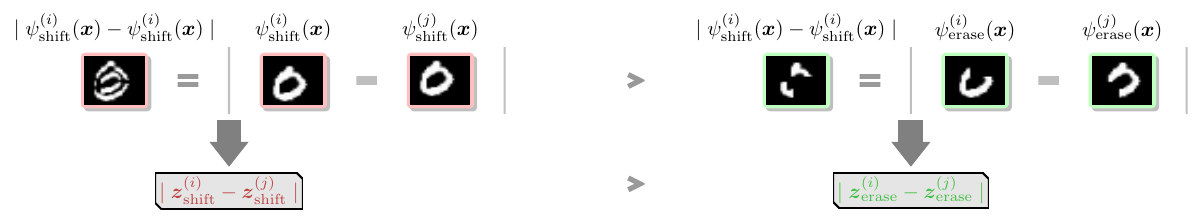}
  \caption{\textbf{Varying Effects of Data Augmentation Methods on the Variance of Logits.} Under Hölder continuity, larger changes in the input lead to larger changes in the logits. \textbf{Left:} Translation preserves the semantic content of the image but introduces large pixel-wise differences, resulting in a higher variations in the logits. \textbf{Right:} Random erasing applies stronger modifications to the image, but the overall pixel-wise differences are smaller, leading to more concentrated logits that are better suited to stable aggregation.}
  \label{fig:holder_condition}
\end{figure}

Figure~\ref{fig:holder_condition} provides an intuitive visualization of how different data augmentation strategies influence the variance of logits under the Hölder continuity assumption. According to Eq.~(\ref{eq:holder_condition}), the distance between logit vectors is bounded by the distance between augmented inputs. As illustrated in the left example, translation preserves the semantic structure of the image but introduces large pixel-wise differences between augmented samples, leading to a larger input distance \(\mid\psi_i(\boldsymbol{x}) - \psi_j(\boldsymbol{x})\mid\) and consequently higher variance in the resulting logits. In contrast, the right example shows random erasing, which applies stronger augmentations but gives smaller overall pixel-wise differences. This reduces the variance of the logit vector, producing more concentrated predictions that are better suited for stable aggregation. The figure thus highlights that not all augmentations contribute equally to logit stability. Our proposed augmentation policies (Var = 0.12) exhibit higher variance than random horizontal flipping (Var = 0.005) and random cropping (Var = 0.01), but significantly lower than the composition of raw MixUp and CutMix (Var = 0.16).

\section{Comparison Experiments of StableTTA-I}
\label{app:Comparison Experiments of StableTTA-I}

Although existing studies primarily rely on simple augmentation policies such as flipping and cropping, we consider the sequential sampling and explore a broader range of strategies to better understand their impact on standard TTA performance and provide a more comprehensive evaluation. In Table~\ref{tab:survey_ours}, we conduct experiments on both large and small models, including AlexNet, ResNet (with current best recipes), and ViT\_B\_16 with diverse \(N\), as well as MNASNet0\_5, MobileNetV3-Small, and ShuffleNetV2\_x0\_5. All pretrained models are obtained from TorchVision. The results show that logit averaging slightly outperforms both hard and soft voting aggregation strategies. We adopt standard data augmentation policies with TorchVision default hyperparameters, including horizontal flip (\(\boldsymbol{p}=0.5\)), random crop (padding=32), random affine transformations (maximum rotation=\(15^\circ\), translation=(0.1, 0.1), scale range=(0.9, 1.1), shear=10), random erasing (p=0.1), mixup (\(\alpha=0.2\)), and CutMix (\(\alpha=1.0\)).

\begin{longtable}{c|r|cc|cc|cc}
\caption{Comparison of aggregation strategies (hard voting, soft voting, and logit averaging) under random data augmentation across different models. All results are derived from our aforementioned experiments using TorchVision pretrained models.}
\label{tab:survey_ours} \\
\toprule
& \textbf{Random Data} & \multicolumn{2}{c|}{\textbf{Hard-vot. (\%)}} & \multicolumn{2}{c|}{\textbf{Soft-vot. (\%)}} & \multicolumn{2}{c}{\textbf{Logit-avg. (\%)}} \\
\textbf{Model} & \textbf{Augmentation} & \textbf{Acc@1} & \textbf{Acc@5} & \textbf{Acc@1} & \textbf{Acc@5} & \textbf{Acc@1} & \textbf{Acc@5} \\
\midrule
\endfirsthead

\toprule
\textbf{Model} & \textbf{Augmentation} & \textbf{Acc@1} & \textbf{Acc@5} & \textbf{Acc@1} & \textbf{Acc@5} & \textbf{Acc@1} & \textbf{Acc@5} \\
\midrule
\endhead

\midrule
\multicolumn{8}{r}{\textit{Continued on next page}} \\
\endfoot

\bottomrule
\endlastfoot

\multirow{9}{*}{\rotatebox{90}{\shortstack{AlexNet\\(\# Params = 61.1 M)\\(7.1 GFLOPS)\\(\(N=10\))}}}
& baseline & 56.522 & 79.066 & 56.522 & 79.066 & 56.522 & 79.066 \\
& flip & 56.630 & 56.760 & 57.306 & 79.674 & 57.322 & 79.714 \\
& crop & 55.016 & 55.158 & 55.460 & 78.240 & 55.002 & 78.082 \\ 
& affine & 51.176 & 51.342 & 51.966 & 75.262 & 51.116 & 74.564 \\ 
& erasing & 56.552 & 56.688 & 56.574 & 78.914 & 56.508 & 79.070 \\ 
& mixup & 59.374 & 59.506 & 68.178 & 89.594 & 73.578 & 91.156 \\ 
& CutMix & 63.356 & 63.460 & 65.754 & 87.124 & 66.814 & 87.172 \\
& flip+crop & 55.292 & 55.434 & 55.850 & 78.376 & 55.606 & 78.278 \\ 
& mixup+CutMix & 64.320 & 64.420 & 69.552 & 89.904 & 72.466 & 90.802 \\ \midrule

\multirow{7}{*}{\rotatebox{90}{\shortstack{ResNet50\textsuperscript{\textdagger}\\\footnotesize{(\# Params = 25.6 M)}\\(40.9 GFLOPS)\\(\(N=10\))}}} 
& baseline & 80.858 & 95.434 & 80.858 & 95.434 & 80.858 & 95.434 \\
& flip & 80.740 & 80.774 & 81.048 & 95.586 & 81.040 & 95.572 \\ 
& crop & 80.896 & 80.932 & 80.984 & 95.590 & 80.946 & 95.542 \\ 
& affine & 76.304 & 76.380 & 76.886 & 93.464 & 76.664 & 93.174 \\ 
& erasing & 80.844 & 80.878 & 80.828 & 95.428 & 80.804 & 95.442 \\ 
& mixup & 82.904 & 82.936 & 87.380 & 98.958 & 89.284 & 98.806 \\ 
& CutMix & 89.550 & 89.574 & 90.610 & 99.182 & 90.466 & 99.154 \\
& flip+crop & 81.050 & 81.090 & 81.152 & 95.616 & 81.070 & 95.588 \\ 
& mixup+CutMix & 87.906 & 87.930 & 89.812 & 99.134 & 90.338 & 99.172 \\ \midrule

\multirow{9}{*}{\rotatebox{90}{\shortstack{ViT\_B\_16\\(\# Params = 86.6 M)\\(175.6 GFLOPS)\\(\(N=10\))}}} 
& baseline & 81.072 & 95.318 & 81.072 & 95.318 & 81.072 & 95.318 \\
& flip & 81.034 & 81.086 & 81.024 & 95.302 & 81.024 & 95.318 \\
& crop & 81.258 & 81.300 & 81.422 & 95.556 & 81.188 & 95.526 \\
& affine & 80.134 & 80.186 & 80.250 & 94.880 & 80.116 & 94.904 \\
& erasing & 81.070 & 81.120 & 81.048 & 95.280 & 81.060 & 95.306 \\
& mixup & 83.258 & 83.300 & 86.998 & 98.858 & 89.508 & 98.910 \\
& CutMix & 87.392 & 87.420 & 88.132 & 98.748 & 88.622 & 98.778 \\
& flip+crop & 81.338 & 81.386 & 81.280 & 95.550 & 81.218 & 95.482 \\
& mixup+CutMix & 86.840 & 86.868 & 88.504 & 98.936 & 89.798 & 99.080 \\ \midrule

\multirow{9}{*}{\rotatebox{90}{\shortstack{AlexNet\\(\# Params = 61.1 M)\\(22.75 GFLOPS)\\(\(N=32\))}}} 
& baseline & 56.522 & 79.066 & 56.522 & 79.066 & 56.522 & 79.066 \\
& flip   & 56.520 & 56.652 & 57.322 & 79.722 & 57.404 & 79.736 \\
& crop    & 55.530 & 55.672 & 55.734 & 78.444 & 55.458 & 78.194 \\
& affine  & 52.142 & 52.296 & 52.654 & 75.716 & 51.632 & 75.162 \\
& erasing & 56.552 & 56.688 & 56.540 & 79.036 & 56.550 & 79.094 \\ 
& mixup   & 61.918 & 62.022 & 69.878 & 90.004 & 75.500 & 92.426 \\ 
& CutMix  & 65.854 & 65.960 & 67.672 & 88.374 & 68.062 & 88.122 \\ 
& flip+crop   & 55.950 & 56.096 & 56.316 & 78.728 & 55.840 & 78.526 \\ 
& mixup+CutMix & 67.166 & 67.252 & 71.296 & 90.930 & 74.522 & 91.972 \\ \midrule

\multirow{9}{*}{\rotatebox{90}{\shortstack{ResNet50\textsuperscript{\textdagger}\\(\# Params = 25.6 M)\\(130.9 GFLOPS)\\(\(N=32\))}}} 
& baseline & 80.858 & 95.434 & 80.858 & 95.434 & 80.858 & 95.434 \\
& flip   & 80.842 & 80.874 & 81.012 & 95.598 & 81.072 & 95.578 \\ 
& crop    & 81.058 & 81.098 & 81.122 & 95.626 & 81.072 & 95.582 \\ 
& affine  & 76.762 & 76.844 & 77.222 & 93.542 & 76.818 & 93.342 \\ 
& erasing & 80.844 & 80.878 & 80.832 & 95.414 & 80.838 & 95.406 \\ 
& mixup   & 84.286 & 84.318 & 88.212 & 99.038 & 90.472 & 99.186 \\ 
& CutMix  & 90.394 & 90.410 & 90.964 & 99.248 & 90.770 & 99.228 \\ 
& flip+crop   & 81.236 & 81.274 & 81.186 & 95.654 & 81.196 & 95.600 \\ 
& mixup+CutMix & 89.702 & 89.718 & 90.602 & 99.216 & 91.296 & 99.308 \\ \midrule

\multirow{9}{*}{\rotatebox{90}{\shortstack{ViT\_B\_16\\(\# Params = 86.6 M)\\(561.9 GFLOPS)\\(\(N=32\))}}} 
& baseline & 81.072 & 95.318 & 81.072 & 95.318 & 81.072 & 95.318 \\
& flip   & 81.004 & 81.054 & 81.022 & 95.300 & 81.006 & 95.300 \\ 
& crop    & 81.430 & 81.472 & 81.428 & 95.598 & 81.376 & 95.588 \\ 
& affine  & 80.316 & 80.368 & 80.394 & 95.018 & 80.288 & 95.014 \\ 
& erasing & 81.068 & 81.118 & 81.066 & 95.276 & 81.064 & 95.302 \\ 
& mixup   & 84.976 & 84.914 & 88.176 & 98.966 & 90.566 & 99.292 \\ 
& CutMix  & 87.996 & 88.024 & 88.574 & 98.930 & 89.040 & 98.912 \\ 
& flip+crop   & 81.346 & 81.388 & 81.448 & 95.582 & 81.408 & 95.576 \\ 
& mixup+CutMix & 88.062 & 88.086 & 89.390 & 99.078 & 90.826 & 99.246 \\ \midrule 

\multirow{9}{*}{\rotatebox{90}{\shortstack{MNASNet0\_5\\(\# Params = 2.2 M)\\(1.6 GFLOPS)\\(\(N=16\))}}}
& baseline     & 67.734 & 87.49  & 67.734 & 87.49  & 67.734 & 87.49  \\
& flip         & 67.690 & 67.766 & 68.026 & 87.632 & 68.014 & 87.656 \\ 
& crop         & 66.766 & 66.852 & 67.082 & 86.796 & 66.818 & 86.746 \\ 
& affine       & 60.924 & 61.028 & 61.900 & 83.086 & 60.600 & 82.004 \\ 
& erasing      & 67.768 & 67.846 & 67.776 & 87.470 & 67.706 & 87.414 \\ 
& mixup        & 71.832 & 71.886 & 78.968 & 95.402 & 82.664 & 95.974 \\ 
& CutMix       & 77.238 & 77.284 & 78.938 & 94.724 & 79.078 & 94.514 \\ 
& flip+crop    & 66.844 & 66.926 & 67.032 & 86.918 & 66.964 & 86.866 \\ 
& mixup+CutMix & 77.254 & 77.298 & 80.418 & 95.804 & 82.468 & 96.014 \\ \midrule

\multirow{6}{*}{\rotatebox{90}{\shortstack{\tiny{MobileNet\_V3\_Samll}\\\footnotesize{(\# Params = 2.5 M)}\\(0.96 GFLOPS)\\(\(N=16\))}}}
& baseline     & 67.668 & 87.402 & 67.668 & 87.402 & 67.668 & 87.402 \\
& flip         & 67.646 & 67.720 & 67.960 & 87.668 & 67.880 & 87.648 \\ 
& crop         & 67.882 & 67.958 & 68.236 & 87.714 & 68.048 & 87.590 \\ 
& affine       & 65.542 & 65.632 & 66.030 & 85.966 & 65.552 & 85.990 \\ 
& erasing      & 67.674 & 67.748 & 67.720 & 87.422 & 67.612 & 87.416 \\ 
& mixup        & 71.734 & 71.770 & 78.650 & 95.072 & 82.960 & 96.070 \\ 
& CutMix       & 75.028 & 75.080 & 76.878 & 93.684 & 77.350 & 93.730 \\ 
& flip+crop    & 68.016 & 68.090 & 68.156 & 87.736 & 68.078 & 87.714 \\ 
& mixup+CutMix & 76.498 & 76.552 & 79.702 & 95.346 & 82.330 & 95.966 \\ \midrule

\multirow{9}{*}{\rotatebox{90}{\shortstack{ShuffleNet\_V2\_X0\_5\\(\# Params = 1.4 M)\\(0.64 GFLOPS)\\(\(N=16\))}}}
& baseline     & 60.552 & 81.746 & 60.552 & 81.746 & 60.552 & 81.746 \\
& flip         & 60.448 & 60.546 & 61.376 & 82.352 & 61.394 & 82.470 \\ 
& crop         & 60.128 & 60.230 & 60.528 & 81.516 & 60.372 & 81.504 \\ 
& affine       & 54.350 & 54.474 & 55.018 & 76.736 & 53.604 & 75.894 \\ 
& erasing      & 60.562 & 60.666 & 60.552 & 81.552 & 60.514 & 81.786 \\ 
& mixup        & 65.274 & 65.338 & 72.124 & 91.592 & 77.786 & 93.506 \\ 
& CutMix       & 71.856 & 71.922 & 73.696 & 91.746 & 74.222 & 91.622 \\ 
& flip+crop    & 60.420 & 60.512 & 60.762 & 81.652 & 60.554 & 81.586 \\ 
& mixup+CutMix & 71.116 & 71.178 & 74.708 & 92.590 & 77.520 & 93.472 \\ \midrule 

\multirow{9}{*}{\rotatebox{90}{\shortstack{MNASNet0\_5\\(\# Params = 2.2 M)\\(3.2 GFLOPS)\\(\(N=32\))}}}
& baseline     & 67.734 & 87.49  & 67.734 & 87.49  & 67.734 & 87.49  \\
& flip         & 67.644 & 67.718 & 68.050 & 87.654 & 68.024 & 87.670 \\ 
& crop         & 66.774 & 66.858 & 67.110 & 86.960 & 66.892 & 86.906 \\ 
& affine       & 61.386 & 61.498 & 62.024 & 83.178 & 60.714 & 82.118 \\ 
& erasing      & 67.768 & 67.846 & 67.742 & 87.482 & 67.726 & 87.472 \\ 
& mixup        & 73.166 & 73.218 & 79.676 & 95.556 & 83.550 & 96.384 \\ 
& CutMix       & 78.092 & 78.140 & 79.554 & 95.260 & 79.590 & 94.796 \\ 
& flip+crop    & 66.954 & 67.034 & 67.198 & 86.978 & 66.990 & 86.860 \\ 
& mixup+CutMix & 78.588 & 78.628 & 81.148 & 96.036 & 83.122 & 96.350 \\ \midrule

\multirow{9}{*}{\rotatebox{90}{\shortstack{MobileNet\_V3\_Samll\\(\# Params = 2.5 M)\\(1.92 GFLOPS)\\(\(N=32\))}}}
& baseline     & 67.668 & 87.402 & 67.668 & 87.402 & 67.668 & 87.402 \\
& flip         & 67.584 & 67.658 & 57.924 & 87.670 & 67.920 & 87.660 \\ 
& crop         & 68.050 & 68.132 & 68.288 & 87.758 & 68.130 & 87.686 \\ 
& affine       & 65.762 & 65.836 & 66.104 & 86.090 & 65.668 & 86.032 \\ 
& erasing      & 67.670 & 67.744 & 67.684 & 87.454 & 67.662 & 87.408 \\ 
& mixup        & 72.956 & 72.992 & 79.442 & 95.208 & 83.624 & 96.348 \\ 
& CutMix       & 76.050 & 76.106 & 77.502 & 94.052 & 77.862 & 93.978 \\ 
& flip+crop    & 68.158 & 68.234 & 68.302 & 87.816 & 68.168 & 87.664 \\ 
& mixup+CutMix & 77.768 & 77.802 & 80.496 & 95.578 & 82.964 & 96.310 \\ \midrule

\multirow{9}{*}{\rotatebox{90}{\shortstack{ShuffleNet\_V2\_X0\_5\\(\# Params = 1.4 M)\\(1.28 GFLOPS)\\(\(N=32\))}}}
& baseline     & 60.552 & 81.746 & 60.552 & 81.746 & 60.552 & 81.746 \\
& flip         & 69.564 & 60.668 & 61.354 & 82.412 & 61.458 & 82.482 \\ 
& crop         & 60.480 & 60.576 & 60.756 & 81.622 & 60.526 & 81.660 \\ 
& affine       & 54.804 & 54.922 & 55.306 & 76.920 & 53.848 & 76.146 \\ 
& erasing      & 60.562 & 60.666 & 60.594 & 81.578 & 60.568 & 81.728 \\ 
& mixup        & 66.750 & 66.820 & 73.140 & 91.724 & 78.454 & 93.776 \\ 
& CutMix       & 72.976 & 73.038 & 74.362 & 92.272 & 74.780 & 92.204 \\ 
& flip+crop    & 60.628 & 60.734 & 60.866 & 81.694 & 60.746 & 81.830 \\ 
& mixup+CutMix & 72.634 & 72.678 & 75.392 & 92.970 & 78.270 & 93.934 \\

\end{longtable}

Table~\ref{tab:main_results} reports the detailed StableTTA-I performance shown in Fig.~\ref{fig:main_results} under \(N=32\), where the ``Peak GFLOPs'' column denotes the computational cost of a single forward pass on a \(224 \times 224\) image.

\begin{longtable}{lrr|ll|ll}
\caption{Comparison of base models and the StableTTA-I-enhanced models on the ImageNet-1K dataset: model size, computational cost, and top-1 and top-5 validation accuracy.}
\label{tab:main_results} \\
\toprule
& & \textbf{Peak} & \multicolumn{2}{c|}{\textbf{Baseline (\%)}} & \multicolumn{2}{c}{\textbf{StableTTA-I (\%)}} \\
\textbf{Weight} & \textbf{Params} & \textbf{GFLOPS} & \textbf{Acc@1} & \textbf{Acc@5} & \textbf{Acc@1} & \textbf{Acc@5} \\
\midrule
\endfirsthead

\toprule
\textbf{Weight} & \textbf{Params} & \textbf{GFLOPS} & \textbf{Acc@1} & \textbf{Acc@5} & \textbf{Acc@1} & \textbf{Acc@5} \\
\midrule
\endhead

\midrule
\multicolumn{7}{r}{\textit{Continued on next page}} \\
\endfoot

\bottomrule
\endlastfoot

\rowcolor{gray!25!white}
AlexNet \cite{krizhevsky2012imagenet,torchvision2016} & 61.1M & 0.71 & 56.522 & 79.066 & 89.332 & 98.028 \\

ConvNeXt\_Tiny \cite{liu2022convnet,torchvision2016} & 28.6M & 4.46 & 82.52 & 96.146 & 95.958 & 99.884 \\ \rowcolor{gray!25!white}
ConvNeXt\_Small \cite{liu2022convnet,torchvision2016} & 50.2M & 8.68 & 83.616 & 96.65 & 95.932 & 99.914 \\
ConvNeXt\_Base \cite{liu2022convnet,torchvision2016} & 88.6M & 15.36 & 84.062 & 96.87 & 95.964 & 99.906 \\ \rowcolor{gray!25!white}
ConvNeXt\_Large \cite{liu2022convnet,torchvision2016} & 197.8M & 34.36 & 84.414 & 96.976 & 95.998 & 99.932 \\

DenseNet121 \cite{huang2017densely,torchvision2016} & 8.0M & 2.83 & 74.434 & 91.972 & 94.160 & 99.370 \\ \rowcolor{gray!25!white}
DenseNet161 \cite{huang2017densely,torchvision2016} & 28.7M & 7.73 & 77.138 & 93.56 & 94.814 & 99.572 \\ 
DenseNet169 \cite{huang2017densely,torchvision2016} & 14.1M & 3.36 & 75.6 & 92.806 & 94.600 & 99.510 \\ \rowcolor{gray!25!white}
DenseNet201 \cite{huang2017densely,torchvision2016} & 20.0M & 4.29 & 76.896 & 93.37 & 94.734 & 99.562 \\

EfficientNet\_B0 \cite{tan2019efficientnet,torchvision2016} & 5.3M & 0.39 & 77.692 & 93.532 & 94.988 & 99.678 \\ \rowcolor{gray!25!white}
EfficientNet\_B1 \cite{tan2019efficientnet,torchvision2016} & 7.8M & 0.69 & 78.642 & 94.186 & 94.940 & 99.738 \\
EfficientNet\_B2 \cite{tan2019efficientnet,torchvision2016} & 9.1M & 1.09 & 80.608 & 95.31 & 95.352 & 99.796 \\ \rowcolor{gray!25!white}
EfficientNet\_B3 \cite{tan2019efficientnet,torchvision2016} & 12.2M & 1.83 & 82.008 & 96.054 & 95.816 & 99.862 \\
EfficientNet\_B4 \cite{tan2019efficientnet,torchvision2016} & 19.3M & 4.39 & 83.384 & 96.594 & 95.386 & 99.874 \\ \rowcolor{gray!25!white}
EfficientNet\_B5 \cite{tan2019efficientnet,torchvision2016} & 30.4M & 10.27 & 83.444 & 96.628 & 95.328 & 99.872 \\
EfficientNet\_B6 \cite{tan2019efficientnet,torchvision2016} & 43.0M & 19.07 & 84.008 & 96.916 & 95.504 & 99.892 \\ \rowcolor{gray!25!white}
EfficientNet\_B7 \cite{tan2019efficientnet,torchvision2016} & 66.3M & 37.75 & 84.122 & 96.908 & 93.452 & 99.718 \\

EfficientNet\_V2\_S \cite{tan2021efficientnetv2,torchvision2016} & 21.5M & 8.37 & 84.228 & 96.878 & 96.046 & 99.926 \\ \rowcolor{gray!25!white}  
EfficientNet\_V2\_M \cite{tan2021efficientnetv2,torchvision2016} & 54.1M & 24.58 & 85.112 & 97.156 & 96.102 & 99.920 \\ 
EfficientNet\_V2\_L \cite{tan2021efficientnetv2,torchvision2016} & 118.5M & 56.08 & 85.808 & 97.788 & 96.164 & 99.892 \\ \rowcolor{gray!25!white}

GoogLeNet \cite{szegedy2015going,torchvision2016} & 6.6M & 1.5 & 69.778 & 89.53 & 93.510 & 99.346 \\

Inception\_V3 \cite{szegedy2016rethinking,torchvision2016} & 27.2M & 5.71 & 77.294 & 93.45 & 94.906 & 99.734 \\ \rowcolor{gray!25!white}

MNASNet0\_5 \cite{tan2019mnasnet,torchvision2016} & 2.2M & 0.1 & 67.734 & 87.49 & 93.122 & 99.188 \\
MNASNet0\_75 \cite{tan2019mnasnet,torchvision2016} & 3.2M & 0.21 & 71.18 & 90.496 & 93.868 & 99.484 \\ \rowcolor{gray!25!white}
MNASNet1\_0 \cite{tan2019mnasnet,torchvision2016} & 4.4M & 0.31 & 73.456 & 91.51 & 94.230 & 99.342 \\
MNASNet1\_3 \cite{tan2019mnasnet,torchvision2016} & 6.3M & 0.53 & 76.506 & 93.522 & 94.876 & 99.728 \\ \rowcolor{gray!25!white}

MaxVit\_T \cite{tu2022maxvit,torchvision2016} & 30.9M & 5.56 & 83.7 & 96.722 & 95.996 & 99.902 \\

MobileNet\_V2 \cite{sandler2018mobilenetv2,torchvision2016} & 3.5M & 0.3 & 71.878 & 90.286 & 93.784 & 99.262 \\ \rowcolor{gray!25!white}
MobileNet\_V2\textsuperscript{\textdagger} \cite{sandler2018mobilenetv2,torchvision2021sota} & 3.5M & 0.3 & 72.154 & 90.822 & 93.922 & 99.516 \\

MobileNet\_V3\_Small \cite{howard2019searching,torchvision2016} & 2.5M & 0.06 & 67.668 & 87.402 & 92.830 & 99.094 \\ \rowcolor{gray!25!white}
MobileNet\_V3\_Large \cite{howard2019searching,torchvision2016} & 5.5M & 0.22 & 74.042 & 91.34 & 94.350 & 99.364 \\ 
MobileNet\_V3\_Large\textsuperscript{\textdagger} \cite{howard2019searching,torchvision2021sota} & 5.5M & 0.22 & 75.274 & 92.566 & 94.648 & 99.656 \\ \rowcolor{gray!25!white}

ResNeXt50\_32x4d \cite{xie2017aggregated,torchvision2016} & 25.0M & 4.23 & 77.618 & 93.698 & 95.126 & 99.680 \\
ResNeXt50\_32x4d\textsuperscript{\textdagger} \cite{xie2017aggregated,torchvision2021sota} & 25.0M & 4.23 & 81.198 & 95.34 & 95.978 & 99.916 \\ \rowcolor{gray!25!white}
ResNeXt101\_32x8d \cite{xie2017aggregated,torchvision2016} & 88.8M & 16.41 & 79.312 & 94.526 & 94.862 & 99.648 \\
ResNeXt101\_32x8d\textsuperscript{\textdagger} \cite{xie2017aggregated,torchvision2021sota} & 88.8M & 16.41 & 82.834 & 96.228 & 95.764 & 99.874 \\ \rowcolor{gray!25!white}

ResNet18 \cite{he2016deep,torchvision2016} & 11.7M & 1.81 & 69.758 & 89.078 & 93.176 & 99.194 \\ 
ResNet34 \cite{he2016deep,torchvision2016} & 21.8M & 3.66 & 73.314 & 91.42 & 94.092 & 99.406 \\ \rowcolor{gray!25!white}
ResNet50 \cite{he2016deep,torchvision2016} & 25.6M & 4.09 & 76.13 & 92.862 & 94.518 & 99.496 \\ 
ResNet50\textsuperscript{\textdagger} \cite{he2016deep,torchvision2021sota} & 25.6M & 4.09 & 80.858 & 95.434 & 95.666 & 99.866 \\ \rowcolor{gray!25!white}
ResNet101 \cite{he2016deep,torchvision2016} & 44.5M & 7.8 & 77.374 & 93.546 & 94.858 & 99.540 \\ 
ResNet101\textsuperscript{\textdagger} \cite{he2016deep,torchvision2021sota} & 44.5M & 7.8 & 81.886 & 95.78 & 95.804 & 99.890 \\ \rowcolor{gray!25!white}
ResNet152 \cite{he2016deep,torchvision2016} & 60.2M & 11.51 & 78.312 & 94.046 & 95.018 & 99.558 \\ 
ResNet152\textsuperscript{\textdagger} \cite{he2016deep,torchvision2021sota} & 60.2M & 11.51 & 82.284 & 96.002 & 95.942 & 99.890 \\ \rowcolor{gray!25!white}

ShuffleNet\_V2\_X0\_5 \cite{ma2018shufflenet,torchvision2016} & 1.4M & 0.04 & 60.552 & 81.746 & 91.372 & 98.724 \\
ShuffleNet\_V2\_X1\_0 \cite{ma2018shufflenet,torchvision2016} & 2.3M & 0.14 & 69.362 & 88.316 & 93.380 & 99.342 \\ \rowcolor{gray!25!white}
ShuffleNet\_V2\_X1\_5 \cite{ma2018shufflenet,torchvision2016} & 3.5M & 0.3 & 72.996 & 91.086 & 94.238 & 99.580 \\
ShuffleNet\_V2\_X2\_0 \cite{ma2018shufflenet,torchvision2016} & 7.4M & 0.58 & 76.23 & 93.006 & 94.996 & 99.716 \\ \rowcolor{gray!25!white}

SqueezeNet1\_0 \cite{iandola2016squeezenet,torchvision2016} & 1.2M & 0.82 & 58.092 & 80.42 & 89.030 & 97.878 \\ 
SqueezeNet1\_1 \cite{iandola2016squeezenet,torchvision2016} & 1.2M & 0.35 & 58.178 & 80.624 & 89.642 & 98.020 \\ \rowcolor{gray!25!white}

Swin\_T \cite{liu2021swin,torchvision2016} & 28.3M & 4.49 & 81.474 & 95.776 & 95.798 & 99.864 \\
Swin\_S \cite{liu2021swin,torchvision2016} & 49.6M & 8.74 & 83.196 & 96.36 & 95.894 & 99.914 \\ \rowcolor{gray!25!white}
Swin\_B \cite{liu2021swin,torchvision2016} & 87.8M & 15.43 & 83.582 & 96.64 & 95.832 & 99.916 \\

Swin\_V2\_T \cite{liu2022swin,torchvision2016} & 28.4M & 5.94 & 82.072 & 96.132 & 95.798 & 99.864 \\ \rowcolor{gray!25!white}
Swin\_V2\_S \cite{liu2022swin,torchvision2016} & 49.7M & 11.55 & 83.712 & 96.816 & 96.024 & 99.902 \\ 
Swin\_V2\_B \cite{liu2022swin,torchvision2016} & 87.9M & 20.32 & 84.112 & 96.864 & 95.890 & 99.938 \\ \rowcolor{gray!25!white}

VGG11\_BN \cite{simonyan2014very,torchvision2016} & 132.9M & 7.61 & 70.37 & 89.81 & 93.542 & 99.124 \\
VGG11 \cite{simonyan2014very,torchvision2016} & 132.9M & 7.61 & 69.02 & 88.628 & 93.236 & 99.038 \\ \rowcolor{gray!25!white}
VGG13\_BN \cite{simonyan2014very,torchvision2016} & 133.1M & 11.31 & 71.586 & 90.374 & 93.666 & 99.190 \\
VGG13 \cite{simonyan2014very,torchvision2016} & 133.0M & 11.31 & 69.928 & 89.246 & 93.400 & 99.084 \\ \rowcolor{gray!25!white}
VGG16\_BN \cite{simonyan2014very,torchvision2016} & 138.4M & 15.47 & 73.36 & 91.516 & 93.978 & 98.242 \\
VGG16 \cite{simonyan2014very,torchvision2016} & 138.4M & 15.47 & 71.592 & 90.382 & 93.542 & 99.164 \\ \rowcolor{gray!25!white}
VGG19\_BN \cite{simonyan2014very,torchvision2016} & 143.7M & 19.63 & 74.218 & 91.842 & 94.122 & 99.266 \\
VGG19 \cite{simonyan2014very,torchvision2016} & 143.7M & 19.63 & 72.376 & 90.876 & 93.868 & 99.246 \\ \rowcolor{gray!25!white}

ViT\_B\_16 \cite{dosovitskiy2020image,torchvision2016} & 86.6M & 17.56 & 81.072 & 95.318 & 95.850 & 99.886 \\ 
ViT\_B\_32 \cite{dosovitskiy2020image,torchvision2016} & 88.2M & 4.41 & 75.912 & 92.466 & 95.000 & 99.778 \\ \rowcolor{gray!25!white}
ViT\_L\_16 \cite{dosovitskiy2020image,torchvision2016} & 304.3M & 61.55 & 79.662 & 94.638 & 95.314 & 99.860 \\ 
ViT\_L\_32 \cite{dosovitskiy2020image,torchvision2016} & 306.5M & 15.38 & 76.972 & 93.07 & 95.132 & 99.804 \\ \rowcolor{gray!25!white}

Wide\_ResNet101\_2 \cite{zagoruyko2016wide,torchvision2016} & 126.9M & 22.75 & 78.848 & 94.284 & 95.146 & 99.666 \\
Wide\_ResNet101\_2\textsuperscript{\textdagger} \cite{zagoruyko2016wide,torchvision2021sota} & 126.9M & 22.75 & 82.51 & 96.02 & 95.914 & 99.898 \\ \rowcolor{gray!25!white}
Wide\_ResNet50\_2 \cite{zagoruyko2016wide,torchvision2016} & 68.9M & 11.4 & 78.468 & 94.086 & 95.168 & 99.654 \\
Wide\_ResNet50\_2\textsuperscript{\textdagger} \cite{zagoruyko2016wide,torchvision2021sota} & 68.9M & 11.4 & 81.602 & 95.758 & 95.736 & 99.868 \\ 

\end{longtable}

\section{Ablation Study of StableTTA-I}
\label{app:Ablation Study of StableTTA-I}

\begin{table}[hb]
\caption{Ablation studies of StableTTA-I versus TTA under varying numbers of experts (\(N\)) across multiple architectures. StableTTA-I consistently outperforms TTA under mixup and CutMix.}
\label{tab:ablation_study} 
\centering
\begin{tabular}{c|rr|cc|cc}
\toprule
& & & \multicolumn{2}{c|}{\textbf{TTA (\%)}} & \multicolumn{2}{c}{\textbf{StableTTA-I (\%)}} \\
\textbf{Model} & $N$ & \textbf{GFLOPS} & \textbf{Acc@1} & \textbf{Acc@5} & \textbf{Acc@1} & \textbf{Acc@5} \\
\midrule

\multirow{4}{*}{\shortstack{AlexNet\\(\# Params = 61.1M)}}
& 4 & 2.84 & 65.590 & 87.022 & 78.750 & 93.434 \\ 
& 8 & 5.68 & 68.604 & 89.432 & 84.810 & 96.216 \\  
& 16 & 11.36 & 70.496 & 90.510 & 87.864 & 97.470 \\ 
& 32 & 22.72 & 71.258 & 90.938 & 89.532 & 98.030 \\ \midrule

\multirow{4}{*}{\shortstack{ResNet50\textsuperscript{\textdagger}\\(\# Params = 25.6M)}}
& 4 & 16.36 & 88.556 & 98.858 & 91.276 & 99.108 \\ 
& 8 & 32.72 & 89.542 & 99.098 & 93.878 & 99.676 \\ 
& 16 & 65.44 & 90.232 & 99.172 & 95.192 & 99.800 \\ 
& 32 & 130.88 & 90.612 & 99.240 & 95.646 & 99.852 \\ \midrule

\multirow{4}{*}{\shortstack{ViT\_B\_16\\(\# Params = 86.6M)}}
& 4 & 70.24 & 87.052 & 98.470 & 91.668 & 99.152 \\ 
& 8 & 140.48 & 88.206 & 98.938 & 94.042 & 99.656 \\ 
& 16 & 280.56 & 89.048 & 99.030 & 95.114 & 99.820 \\ 
& 32 & 561.92 & 89.382 & 99.074 & 95.852 & 99.888 \\ \midrule

\multirow{4}{*}{\shortstack{MNASNet0\_5\\(\# Params = 2.2M)}}
& 4 & 0.4 & 76.806 & 93.714 & 85.396 & 96.614 \\ 
& 8 & 0.8 & 79.358 & 95.276 & 89.856 & 98.214 \\ 
& 16 & 1.6 & 80.682 & 95.894 & 91.882 & 98.864 \\ 
& 32 & 3.2 & 81.162 & 96.030 & 93.210 & 99.222 \\ \midrule

\multirow{4}{*}{\shortstack{MobileNet\_V3\_Samll\\(\# Params = 2.5M)}}
& 4 & 0.24 & 75.716 & 93.094 & 85.192 & 96.392 \\
& 8 & 0.48 & 78.540 & 94.664 & 89.588 & 97.992 \\
& 16 & 0.96 & 79.834 & 95.294 & 91.770 & 98.720 \\
& 32 & 1.92 & 80.552 & 95.602 & 92.870 & 98.998 \\ \midrule

\multirow{4}{*}{\shortstack{ShuffleNet\_V2\_X0\_5\\(\# Params = 1.4M)}}
& 4 & 0.16 & 70.238 & 89.682 & 81.764 & 94.662 \\
& 8 & 0.32 & 73.140 & 91.694 & 86.936 & 97.142 \\
& 16 & 0.64 & 74.536 & 92.532 & 90.052 & 98.358 \\
& 32 & 1.28 & 75.556 & 92.898 & 91.390 & 98.704\\

\bottomrule
\end{tabular}
\end{table}

Table~\ref{tab:ablation_study} reports the detailed performance comparison between standard TTA and our StableTTA-I under mixup and CutMix augmentation policies, across different architectures and varying numbers of experts (\(N\)). Consistent with the trends observed in Fig.~\ref{fig:ablation_study}, StableTTA-I outperforms TTA. Larger \(N\) steadily improves both top-1 and top-5 accuracies for all models, including heavyweight models such as AlexNet, ResNet50, and ViT\_B\_16, as well as lightweight models such as MNASNet0\_5, MobileNet\_V3\_Samll, and ShuffleNet\_V2\_X0\_5. However, the performance gains gradually plateau as \(N\) increases, indicating a clear saturation effect. This table provides evidences supporting the advantages of StableTTA.

\section{Sensitivity Analysis of StableTTA-I}
\label{app:Sensitivity Analysis of StableTTA-I}

Table~\ref{tab:sensitivity_analysis} presents a detailed sensitivity analysis of StableTTA-I with respect to the number of candidates \(K\) and experts \(N\) across different architectures. Consistent with the trends observed in Fig.~\ref{fig:sensitivity_analysis}, StableTTA-I exhibits strong robustness to the choices of \(K\), with performance remaining largely stable across a wide range of values. Increasing \(N\) consistently improves both top-1 and top-5 accuracies, while the impact of varying \(K\) is relatively minor. Notably, when logit processing is disabled (when \(K=C\)), performance drops significantly across all models, reaffirming the critical role of our logit processing (NSS) in achieving accuracy gains. Overall, these results demonstrate that StableTTA-I delivers stable, reliable improvements without requiring meticulous hyperparameter tuning.

\begin{longtable}{cl|cc|cc|cc|cc}
\caption{Sensitivity analysis of StableTTA-I over \(K\) and \(N\). Results show robustness to \(K\), consistent gains with larger \(N\), and significant drops when logit processing is disabled (\(K=C\)).}
\label{tab:sensitivity_analysis} \\
\toprule
& 
& \multicolumn{2}{c|}{\(N=4\)} 
& \multicolumn{2}{c|}{\(N=8\)} 
& \multicolumn{2}{c|}{\(N=16\)} 
& \multicolumn{2}{c}{\(N=32\)} \\
\textbf{Model} & \(K\) & \textbf{Acc@1} & \textbf{Acc@5} & \textbf{Acc@1} & \textbf{Acc@5} & \textbf{Acc@1} & \textbf{Acc@5} & \textbf{Acc@1} & \textbf{Acc@5} \\
\midrule
\endfirsthead

\toprule
\textbf{Model} & \(K\) & \textbf{Acc@1} & \textbf{Acc@5} & \textbf{Acc@1} & \textbf{Acc@5} & \textbf{Acc@1} & \textbf{Acc@5} & \textbf{Acc@1} & \textbf{Acc@5} \\
\midrule
\endhead

\midrule
\multicolumn{10}{r}{\textit{Continued on next page}} \\
\endfoot

\bottomrule
\endlastfoot

\multirow{6}{*}{\rotatebox{90}{\shortstack{AlexNet\\(61.1M)\\(0.71GFLOPS)}}} 
& 1  & 78.412 & 93.182 & 84.638 & 96.134 & 87.942 & 97.530 & 89.576 & 98.052 \\
& 2  & 78.444 & 93.262 & 84.982 & 96.312 & 87.946 & 97.528 & 89.496 & 98.052 \\
& 5  & 78.804 & 93.328 & 84.666 & 96.220 & 87.984 & 97.412 & 89.484 & 98.108 \\
& 10 & 78.750 & 93.434 & 84.810 & 96.216 & 87.864 & 97.470 & 89.532 & 98.030 \\
& 20 & 78.540 & 93.320 & 84.778 & 96.160 & 88.072 & 97.540 & 89.530 & 98.076 \\
& C  & 66.212 & 83.420 & 71.606 & 91.158 & 74.948 & 94.520 & 76.918 & 96.222 \\

\midrule
\multirow{6}{*}{\rotatebox{90}{\shortstack{ResNet50\textsuperscript{\textdagger}\\(25.6M)\\(4.09GFLOPS)}}} 
& 1  & 91.524 & 99.082 & 93.788 & 99.614 & 95.076 & 99.802 & 95.706 & 99.860 \\
& 2  & 91.436 & 99.120 & 93.830 & 99.624 & 95.022 & 99.780 & 95.552 & 99.834 \\
& 5  & 91.484 & 99.084 & 93.746 & 99.622 & 94.998 & 99.818 & 95.656 & 99.860 \\
& 10 & 91.276 & 99.108 & 93.878 & 99.676 & 95.192 & 99.800 & 95.646 & 99.852 \\
& 20 & 91.576 & 99.126 & 93.726 & 99.598 & 95.008 & 99.770 & 95.642 & 99.876 \\
& C  & 86.670 & 96.534 & 89.820 & 98.822 & 92.034 & 99.542 & 93.110 & 99.744 \\

\midrule
\multirow{6}{*}{\rotatebox{90}{\shortstack{ViT\_B\_16\\(86.6M)\\(17.6GFLOPS)}}}
& 1  & 91.488 & 99.142 & 93.906 & 99.676 & 95.308 & 99.822 & 95.822 & 99.864 \\
& 2  & 91.604 & 99.128 & 94.000 & 99.670 & 95.304 & 99.810 & 95.826 & 99.878 \\
& 5  & 91.602 & 99.112 & 93.894 & 99.682 & 95.088 & 99.820 & 95.824 & 99.884 \\
& 10 & 91.668 & 99.152 & 94.042 & 99.656 & 95.114 & 99.820 & 95.852 & 99.888 \\
& 20 & 91.468 & 99.126 & 94.972 & 99.648 & 95.234 & 99.816 & 95.940 & 99.904 \\
& C  & 86.710 & 96.578 & 89.898 & 98.844 & 91.648 & 99.540 & 92.832 & 99.744 \\

\midrule
\multirow{6}{*}{\rotatebox{90}{\shortstack{MNASNet0\_5\\(2.2M)\\(0.1GFLOPS)}}} 
& 1  & 85.416 & 96.534 & 89.670 & 98.222 & 92.090 & 98.928 & 93.060 & 99.172 \\
& 2  & 85.380 & 96.450 & 89.880 & 98.250 & 91.900 & 98.898 & 93.156 & 99.190 \\
& 5  & 85.314 & 96.226 & 89.842 & 98.140 & 92.030 & 98.854 & 93.240 & 99.222 \\
& 10 & 85.396 & 96.614 & 89.856 & 98.214 & 91.882 & 98.864 & 93.210 & 99.222 \\
& 20 & 85.344 & 96.500 & 89.868 & 98.226 & 92.124 & 98.934 & 93.088 & 99.178 \\
& C  & 76.678 & 90.734 & 81.654 & 96.026 & 84.640 & 97.886 & 86.572 & 98.760 \\

\midrule
\multirow{6}{*}{\rotatebox{90}{\shortstack{{\scriptsize{MobileNet\_V3\_Small}}\\(2.5M)\\(0.06GFLOPS)}}} 
& 1  & 85.216 & 96.384 & 89.566 & 98.028 & 91.660 & 98.712 & 92.814 & 99.074 \\
& 2  & 85.356 & 96.324 & 89.540 & 97.988 & 91.654 & 98.680 & 92.818 & 99.006 \\
& 5  & 85.510 & 96.480 & 89.440 & 97.994 & 91.762 & 98.730 & 92.746 & 99.014 \\
& 10 & 85.192 & 96.392 & 89.588 & 97.992 & 91.770 & 98.720 & 92.870 & 98.998 \\
& 20 & 85.290 & 96.320 & 89.492 & 98.014 & 91.738 & 98.776 & 92.774 & 98.992 \\
& C  & 76.148 & 90.256 & 81.320 & 95.842 & 84.176 & 97.896 & 86.002 & 98.664 \\

\midrule
\multirow{6}{*}{\rotatebox{90}{\shortstack{{\scriptsize{ShuffleNet\_V2\_X0\_5}}\\(1.4M)\\(0.04GFLOPS)}}} 
& 1  & 81.842 & 94.856 & 87.246 & 97.186 & 89.904 & 98.274 & 91.216 & 98.712 \\
& 2  & 82.136 & 95.046 & 87.038 & 97.178 & 90.088 & 98.302 & 91.326 & 98.700 \\
& 5  & 81.730 & 94.694 & 87.208 & 97.346 & 89.868 & 98.230 & 91.348 & 98.674 \\
& 10 & 81.764 & 94.662 & 86.936 & 97.142 & 90.052 & 98.358 & 91.390 & 98.704 \\
& 20 & 81.492 & 94.622 & 87.230 & 97.314 & 90.116 & 98.240 & 91.426 & 98.756 \\
& C  & 71.392 & 87.140 & 76.288 & 93.456 & 79.464 & 96.174 & 81.690 & 97.606 \\

\end{longtable}

Additionally, when the batch size is set to 32, the batch boundaries and class boundaries repeat every
\[
\operatorname{lcm}(50, 32) = 800,
\]
each cycle contains \(800/32=25\) batches. Within each cycle, the number of samples from the majority class in each batch follows the pattern:
\[
32,18,32,28,22,32,24,26,32,20,30,32,16,32,30,20,32,26,24,32,22,28,32,18,32.
\]
The average majority-class ratio is \(672/800=84\%\). In this case, the model performance of both small models (such as MobileNetV2) and stronger models (e.g., ResNet and ViT) decreases by 0.5–1.2\% in accuracy compared to the case with batch size 16.

If the batch size is set to 2, 5, 10, 25, or 50 (i.e., any divisor of 50), then each batch lies entirely within a single class. Therefore, every batch is class-pure, and the majority-class ratio is 100\%. In this case, even with a batch size of 5 on MobileNetV2, the model accuracy can approach 98\%.

\section{Means and Standard Deviations of Repeated Experimental Evaluation of StableTTA-I}
\label{app:Means and Standard Deviations of Repeated Experimental Evaluation of StableTTA-I}
In Table~\ref{tab:statistical_significance}, we report the mean and standard deviation of top-1 and top-5 accuracy on the ImageNet-1K validation dataset for both heavyweight models (e.g., AlexNet, ResNet50, and ViT\_B\_16) and lightweight models (e.g., MNASNet0\_5, MobileNet\_V3\_Small, and ShuffleNet\_V2\_X0\_5) boosted by our StableTTA. Each experiment is conducted independently and repeated five times. The table shows that the standard deviation ranges from approximately 0.01\% to 0.1\%.

\begin{table}[ht]
\caption{ImageNet-1K validation accuracy (Top-1/Top-5). Mean and standard deviation are reported over five independent runs for both heavyweight and lightweight models.}
\label{tab:statistical_significance} 
\centering
\begin{tabular}{r|cc|cc|cc}
\toprule
& \multicolumn{2}{c|}{\textbf{AlexNet (\%)}} & \multicolumn{2}{c|}{\textbf{ResNet50\textsuperscript{\textdagger} (\%)}} & \multicolumn{2}{c}{\textbf{ViT\_B\_16 (\%)}} \\
& \textbf{Acc@1} & \textbf{Acc@5} & \textbf{Acc@1} & \textbf{Acc@5} & \textbf{Acc@1} & \textbf{Acc@5} \\
\midrule

& 89.492 & 98.034 & 95.596 & 99.862 & 95.888 & 99.868 \\
& 89.558 & 98.056 & 95.622 & 99.854 & 95.830 & 99.898 \\
& 89.642 & 98.020 & 95.620 & 99.868 & 95.916 & 99.896 \\
& 89.620 & 98.070 & 95.684 & 99.842 & 95.848 & 99.882 \\
& 89.588 & 98.096 & 95.666 & 99.858 & 95.876 & 99.878 \\ \cmidrule(lr){2-7}
mean & 89.580 & 98.055 & 95.638 & 99.857 & 95.872 & 99.884 \\
std & 0.059 & 0.030 & 0.036 & 0.010 & 0.034 & 0.013 \\ \midrule
& \multicolumn{2}{c|}{\textbf{MNASNet0\_5}} & \multicolumn{2}{c|}{\textbf{MobileNet\_V3\_Small}} & \multicolumn{2}{c}{\textbf{ShuffleNet\_V2\_X0\_5}} \\
& \textbf{Acc@1} & \textbf{Acc@5} & \textbf{Acc@1} & \textbf{Acc@5} & \textbf{Acc@1} & \textbf{Acc@5} \\
\midrule

& 93.014 & 99.256 & 92.772 & 99.004 & 91.222 & 98.686 \\
& 93.092 & 99.212 & 92.784 & 99.060 & 91.334 & 98.750 \\
& 93.288 & 99.204 & 92.746 & 99.052 & 91.370 & 98.764 \\
& 93.090 & 99.208 & 92.896 & 99.000 & 91.128 & 98.690 \\
& 93.138 & 99.262 & 92.882 & 99.032 & 91.314 & 98.772 \\ \cmidrule(lr){2-7}
mean & 93.124 & 99.228 & 92.816 & 99.030 & 91.274 & 98.732 \\
std & 0.103 & 0.026 & 0.070 & 0.026 & 0.099 & 0.039 \\

\bottomrule
\end{tabular}
\end{table}

\section{Implementation Details of StableTTA-II}
\label{app:StableTTA-II: Implementation Details}

Let \(\boldsymbol{h} \in \mathbb{R}^{C' \times H \times W}\) denote a feature map. The CNN head typically comprises a global average pooling layer followed by a linear layer \cite{liu2022convnet,huang2017densely,tan2019efficientnet,tan2019mnasnet,sandler2018mobilenetv2,howard2019searching,xie2017aggregated,he2016deep,ma2018shufflenet,zagoruyko2016wide}. The overall computational complexity of the global average pooling layer is:
\[
\underbrace{C'(HW-1)}_{\text{additions}} 
\;+\; 
\underbrace{C'}_{\text{mean scaling}} = C'HW.
\]
The overall computational complexity of a linear layer \(\text{Linear}(C', C)\) is:
\[
\underbrace{C'C}_{\text{multiplications}} 
\;+\; 
\underbrace{(C'-1)C}_{\text{adds (dot product)}} 
\;+\; 
\underbrace{C}_{\text{bias adds}}
= 2C'C.
\]

\begin{table}[h]
\caption{Feature-level cropping configurations used in StableTTA-II across different architectures.}
\label{tab:archicture} 
\centering
\renewcommand{\arraystretch}{1.2}
\begin{tabular}{l|p{5.75cm}p{4.5cm}}
\toprule
\textbf{Model} & \textbf{Head} & \textbf{Crops} \\
\midrule

ConvNeXt\_Tiny & AvgPool+Norm+Linear(768,1000) & \(\boldsymbol{h}, \boldsymbol{h}_{1:-1, :}, \boldsymbol{h}_{:-1, :}, \boldsymbol{h}_{1:, :}, \boldsymbol{h}_{:, 1:-1}\) \\
ConvNeXt\_Small & AvgPool+Norm+Linear(768,1000) & \(\boldsymbol{h}, \boldsymbol{h}_{1:-1, :}, \boldsymbol{h}_{:-1, :}, \boldsymbol{h}_{1:, :}, \boldsymbol{h}_{:, 1:-1}\) \\
ConvNeXt\_Base & AvgPool+Norm+Linear(1024,1000) & \(\boldsymbol{h}, \boldsymbol{h}_{1:-1, :}, \boldsymbol{h}_{:-1, :}, \boldsymbol{h}_{1:, :}, \boldsymbol{h}_{:, 1:-1}\) \\
ConvNeXt\_Large & AvgPool+Norm+Linear(1536,1000) & \(\boldsymbol{h}, \boldsymbol{h}_{1:-1, :}, \boldsymbol{h}_{:-1, :}, \boldsymbol{h}_{1:, :}, \boldsymbol{h}_{:, 1:-1}\) \\
\midrule
DenseNet121 & ReLU+AvgPool+Linear(1024,1000) & \(\boldsymbol{h}, \boldsymbol{h}_{1:-1, :}, \boldsymbol{h}_{:, 1:-1}\) \\
DenseNet161 & ReLU+AvgPool+Linear(2208,1000) & \(\boldsymbol{h}, \boldsymbol{h}_{1:-1, :}, \boldsymbol{h}_{:, 1:-1}\) \\
DenseNet169 & ReLU+AvgPool+Linear(1664,1000) & \(\boldsymbol{h}, \boldsymbol{h}_{1:-1, :}, \boldsymbol{h}_{:, 1:-1}\) \\
DenseNet201 & ReLU+AvgPool+Linear(1920,1000) & \(\boldsymbol{h}, \boldsymbol{h}_{1:-1, :}, \boldsymbol{h}_{:, 1:-1}\) \\
\midrule
EfficientNet\_B0 & AvgPool+DropOut+Linear(1280,1000) & \(\boldsymbol{h}, \boldsymbol{h}_{1:-1, :}, \boldsymbol{h}_{:, 1:-1}, \boldsymbol{h}_{1:-1, 1:-1}\) \\
EfficientNet\_B1 & AvgPool+DropOut+Linear(1280,1000) & \(\boldsymbol{h}, \boldsymbol{h}_{1:-1, :}, \boldsymbol{h}_{:, 1:-1}, \boldsymbol{h}_{1:-1, 1:-1}\) \\
EfficientNet\_B2 & AvgPool+DropOut+Linear(1408,1000) & \(\boldsymbol{h}, \boldsymbol{h}_{1:-1, :}, \boldsymbol{h}_{:, 1:-1}, \boldsymbol{h}_{1:-1, 1:-1}\) \\
EfficientNet\_B3 & AvgPool+DropOut+Linear(1536,1000) & \(\boldsymbol{h}, \boldsymbol{h}_{2:-2, :}, \boldsymbol{h}_{:, 2:-2}, \boldsymbol{h}_{2:-2, 2:-2}\) \\
EfficientNet\_B4 & AvgPool+DropOut+Linear(1792,1000) & \(\boldsymbol{h}, \boldsymbol{h}_{2:-2, :}, \boldsymbol{h}_{:, 2:-2}, \boldsymbol{h}_{2:-2, 2:-2}\) \\
\midrule
MNASNet0\_5 & AvgPool+DropOut+Linear(1280,1000) & \(\boldsymbol{h}, \boldsymbol{h}_{1:-1, :}, \boldsymbol{h}_{:, 1:-1}\) \\
MNASNet0\_75 & AvgPool+DropOut+Linear(1280,1000) & \(\boldsymbol{h}, \boldsymbol{h}_{1:-1, :}, \boldsymbol{h}_{:, 1:-1}\) \\
MNASNet1\_0 & AvgPool+DropOut+Linear(1280,1000) & \(\boldsymbol{h}, \boldsymbol{h}_{1:-1, :}, \boldsymbol{h}_{:, 1:-1}\) \\
MNASNet1\_3 & AvgPool+DropOut+Linear(1280,1000) & \(\boldsymbol{h}, \boldsymbol{h}_{1:-1, :}, \boldsymbol{h}_{:, 1:-1}\) \\
\midrule
MobileNet\_V2 & AvgPool+DropOut+Linear(1280,1000) & \(\boldsymbol{h}, \boldsymbol{h}_{1:-1, :}, \boldsymbol{h}_{:, 1:-1}, \boldsymbol{h}_{1:-1, 1:-1}\) \\
MobileNet\_V3\_Small & {\tiny AvgPool+Linear(576,1024)+Hardswish+DropOut+Linear(1024,1000)} & \(\boldsymbol{h}, \boldsymbol{h}_{1:-1, :}, \boldsymbol{h}_{:, 1:-1}, \boldsymbol{h}_{1:-1, 1:-1}\) \\
MobileNet\_V3\_Large & {\tiny AvgPool+Linear(960,1280)+Hardswish+DropOut+Linear(1280,1000)} & \(\boldsymbol{h}, \boldsymbol{h}_{1:-1, :}, \boldsymbol{h}_{:, 1:-1}, \boldsymbol{h}_{1:-1, 1:-1}\) \\
\midrule
ResNeXt50\_32x4d & AvgPool+Linear(2048,1000) & \(\boldsymbol{h}, \boldsymbol{h}_{1:-1, :}, \boldsymbol{h}_{:, 1:-1}\) \\
ResNeXt101\_32x8d & AvgPool+Linear(2048,1000) & \(\boldsymbol{h}, \boldsymbol{h}_{1:-1, :}, \boldsymbol{h}_{:, 1:-1}\) \\
ResNeXt101\_64x4d & AvgPool+Linear(2048,1000) & \(\boldsymbol{h}, \boldsymbol{h}_{1:-1, :}, \boldsymbol{h}_{:, 1:-1}\) \\
\midrule
ResNet18 & AvgPool+Linear(512,1000) & \(\boldsymbol{h}, \boldsymbol{h}_{1:-1, :}, \boldsymbol{h}_{:, 1:-1}\) \\
ResNet34 & AvgPool+Linear(512,1000) & \(\boldsymbol{h}, \boldsymbol{h}_{1:-1, :}, \boldsymbol{h}_{:, 1:-1}\) \\
ResNet50 & AvgPool+Linear(2048,1000) & \(\boldsymbol{h}, \boldsymbol{h}_{1:-1, :}, \boldsymbol{h}_{:, 1:-1}\) \\
ResNet101 & AvgPool+Linear(2048,1000) & \(\boldsymbol{h}, \boldsymbol{h}_{1:-1, :}, \boldsymbol{h}_{:, 1:-1}\) \\
ResNet152 & AvgPool+Linear(2048,1000) & \(\boldsymbol{h}, \boldsymbol{h}_{1:-1, :}, \boldsymbol{h}_{:, 1:-1}\) \\
\midrule
ShuffleNet\_V2\_X0\_5 & AvgPool+Linear(1024,1000) & \(\boldsymbol{h}, \boldsymbol{h}_{1:-1, :}, \boldsymbol{h}_{:, 1:-1}, \boldsymbol{h}_{1:-1, 1:-1}\) \\
ShuffleNet\_V2\_X1\_0 & AvgPool+Linear(1024,1000) & \(\boldsymbol{h}, \boldsymbol{h}_{1:-1, :}, \boldsymbol{h}_{:, 1:-1}, \boldsymbol{h}_{1:-1, 1:-1}\) \\
ShuffleNet\_V2\_X1\_5 & AvgPool+Linear(1024,1000) & \(\boldsymbol{h}, \boldsymbol{h}_{1:-1, :}, \boldsymbol{h}_{:, 1:-1}, \boldsymbol{h}_{1:-1, 1:-1}\) \\
ShuffleNet\_V2\_X2\_0 & AvgPool+Linear(1024,1000) & \(\boldsymbol{h}, \boldsymbol{h}_{1:-1, :}, \boldsymbol{h}_{:, 1:-1}, \boldsymbol{h}_{1:-1, 1:-1}\) \\
\midrule
Wide\_ResNet50\_2 & AvgPool+Linear(2048,1000) & \(\boldsymbol{h}, \boldsymbol{h}_{1:-1, :}, \boldsymbol{h}_{:, 1:-1}\) \\
Wide\_ResNet101\_2 & AvgPool+Linear(2048,1000) & \(\boldsymbol{h}, \boldsymbol{h}_{1:-1, :}, \boldsymbol{h}_{:, 1:-1}\) \\

\bottomrule
\end{tabular}
\end{table}

Table~\ref{tab:archicture} illustrates the feature-level cropping configurations used in StableTTA-II across different architectures. In most model architectures, both the feature height \(H\) and width \(W\) are no larger than 9. 
Therefore, we typically choose \(k=1\), which corresponds to the integer that is closest to removing \((1-224/256)\) proportion size along both the horizontal and vertical directions of the input images. 
This choice is motivated by the fact that \(k=1\) is the closest integer such that \(k/9\) approximates \((1-224/256)\).

It is worth noting that StableTTA-II cannot be replaced with weighted features:
\[
\frac{1}{N}\sum_{i=1}^{N}\operatorname{Head}(\operatorname{Crop}^{(i)}(\boldsymbol{h})) \neq \operatorname{Head}(\frac{1}{N}\sum_{i=1}^{N}\operatorname{Crop}^{(i)}(\boldsymbol{h})).
\]

Although some model heads consist primarily of global average pooling followed by a linear layer, the above equation holds if and only if the model head is linear that contains only a global average pooling layer followed by a linear layer without bias or activation functions.

\section{Extra Experiments of StableTTA-II}
\label{app:StableTTA-II: Extra Experiments}
Flipping and 90-degree rotations are not valid transformations at the feature level because CNN classification heads typically begin with a global average pooling layer, which enforces invariance to such operations:
\[
\operatorname{AvgPool}(\psi(\boldsymbol{h})) = \boldsymbol{h}, \quad \psi \in \{\operatorname{Flip}, \operatorname{Rotation}_{90}, \operatorname{Rotation}_{180}, \operatorname{Rotation}_{270} \}.
\]
In this appendix, we evaluate several cropping strategies:

\begin{itemize}
    \item TripleCrop: identity + horizontal center + vertical center
    \item FiveCrop: identity + left + right + top + bottom
    \item SevenCrop: identity + left + right + top + bottom + horizontal center + vertical center
    \item NineCrop: identity + left + right + top + bottom + top-left + top-right + bottom-left + bottom-right
\end{itemize}

We also introduce \(\operatorname{RandomCrop}(p)\), where \(p\) denotes the probability of sampling from the set of eight spatial crops (left, right, top, bottom, and four corners). This sampling is repeated 8 times, and the identity crop is always appended to ensure at least one valid (non-empty) selection.

Additionally, we consider Random Erasing, which randomly masks feature-level pixels with probability \(p\) before feeding them into the model head.

\begin{figure}[h]
  \centering
  \includegraphics[width=\columnwidth]{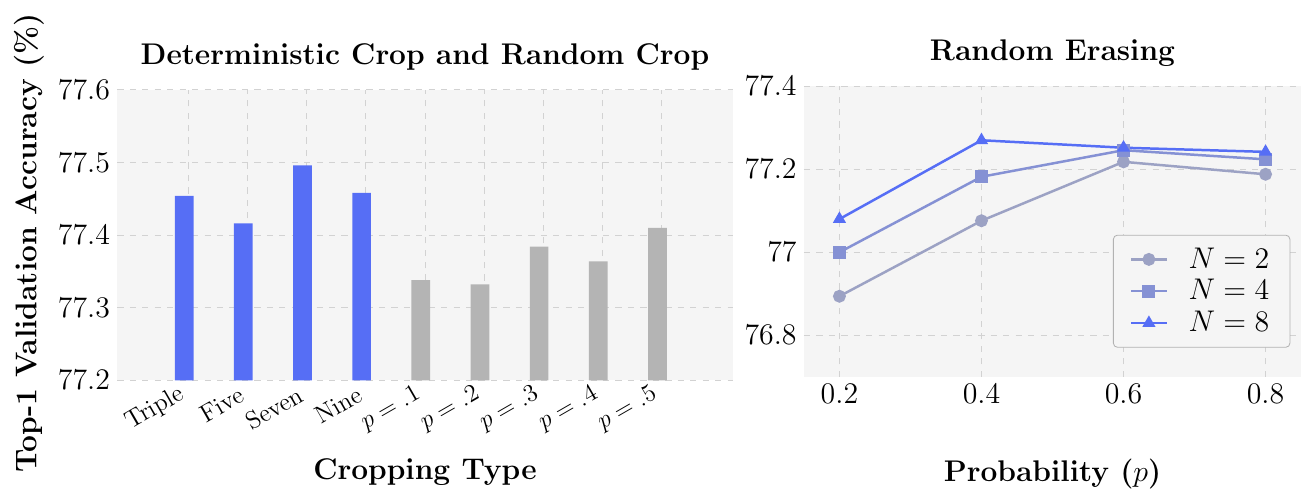}
  \caption{\textbf{Sensitivity analysis of cropping strategies and random erasing.} Left: Comparison of deterministic multi-crop methods (TripleCrop, FiveCrop, SevenCrop, NineCrop) and RandomCrop with varying probabilities \(p\). Right: Effect of Random Erasing probability p on top-1 validation accuracy for different numbers of experts \(N \in \{2,4,8\}\). Overall, moderate cropping diversity improves performance, while excessive randomness provides degraded returns.}
  \label{fig:extra_expriments}
\end{figure}

Fig~\ref{fig:extra_expriments} illustrates the sensitivity of model performance to different cropping strategies and random erasing probabilities.

\section{Additional Discussion on Addressing the Computational Cost of Ensemble Methods}
\label{app:Additional Discussion on Addressing the Computational Cost of Ensemble Methods}
It is worth noting that nearly all CNNs accept dynamically sized input images. Therefore, one possible solution to reduce the computational cost of TTA is to resize the augmented images to a lower resolution. However, this idea does not work well in practice. For example, we randomly apply 10-crop to a raw \(224 \times 224\) image, producing 10 individual images with resize resolutions of \(224 \times 224\), and then resize all of them to \(112 \times 112\) before feeding them into MobileNetV2. In this way, the computational cost of each inference is reduced, but the final validation accuracy drops from 71.878\% to 55.816\%.

Another unintuitive fact is that ImageNet-1K models usually resize images to \(256 \times 256\) and then apply center cropping to obtain \(224 \times 224\) images for standard validation. If we keep the full image and directly resize it to \(224 \times 224\), the validation accuracy becomes worse than the baseline. Both phenomena are easy to verify and reproduce.

Now let us reconsider why people do not split a large image into patches as model inputs and then aggregate the final outputs to reduce the computational cost of each forward pass. For example, we can divide a \(224 \times 224\) image into four \(112 \times 112\) patches and sum the corresponding logits from the four patches. However, this idea also does not work well. Our experiments show that, for MobileNetV2, the final validation accuracy drops from 71.878\% to 62.780\%.



\end{document}